\documentclass[acmtog]{acmart}

\usepackage{booktabs} 
\usepackage{tikz}
\usepackage{tabularx}
\usepackage[normalem]{ulem}
\usepackage{caption} 
\usepackage{pbox} 
\usepackage{mathtools}
\usepackage[ruled]{algorithm2e} 
\usepackage{subcaption}
\usepackage{paralist} 

\definecolor{yellow}{rgb}{0.6, 0.6, 0.2}
\definecolor{darkyellow}{rgb}{0.8, 0.8, 0.5}
\definecolor{orange}{rgb}{1, 0.8, 0.6}
\definecolor{red}{rgb}{1, 0.3, 0.3}
\definecolor{darkred}{rgb}{0.7, 0.3, 0.3}
\definecolor{darkgreen}{rgb}{0.3, 0.7, 0.3}
\definecolor{blue}{rgb}{0, 0, 1.0}
\definecolor{green}{rgb}{0, 1.0, 0}
\definecolor{pink}{rgb}{1, 0.4, 0.7}

\newcommand{\myvec}[1]{\vec{#1}}
\newcommand{\lossfun}[1]{\mathcal{L}_{\mathrm{#1}}}
\newcommand{\light}{\ell}
\newcommand{\norm}[1]{\left\lVert#1\right\rVert}

\newcommand{\neighborhood}{\Omega}
\newcommand{\lightcount}{m}
\newcommand{\fillintensity}{P_{\mathrm{fill}}}
\newcommand{\keyintensity}{P_{\mathrm{key}}}
\newcommand{\myimage}{I}

\DeclareMathOperator*{\select}{select}

\newcommand{\ie}{\textit{i}.\textit{e}., }
\newcommand{\eg}{\textit{e}.\textit{g}. }

\newcommand{\nosv}{No-SV}
\newcommand{\noss}{No-SS}
\newcommand{\nocolor}{No-Color}

\SetAlFnt{\small}
\SetAlCapFnt{\small}
\SetAlCapNameFnt{\small}
\SetAlCapHSkip{0pt}

\newlength{\itemwidth} 
\newcolumntype{Y}{>{\centering\arraybackslash}X} 
\newcolumntype{P}[1]{>{\centering\arraybackslash}p{#1}} 
\usetikzlibrary{calc} 
\usetikzlibrary{tikzmark} 
\usetikzlibrary{spy} 

\acmJournal{TOG}
\acmYear{2020}
\acmVolume{39}
\acmNumber{4}
\acmArticle{78}
\acmMonth{7}

\setcopyright{acmcopyright}

\acmDOI{10.1145/3386569.3392390}

\begin{document}
\title{Portrait Shadow Manipulation}

\author{Xuaner (Cecilia) Zhang}
\affiliation{%
 \institution{University of California, Berkeley}}
\email{cecilia77@berkeley.edu}

\author{Jonathan T. Barron}
\affiliation{%
 \institution{Google Research}}
\email{barron@google.com}

\author{Yun-Ta Tsai}
\affiliation{%
 \institution{Google Research}}
\email{yuntatsai@google.com}

\author{Rohit Pandey}
\affiliation{%
 \institution{Google}}
\email{rohitpandey@google.com}

\author{Xiuming Zhang}
\affiliation{%
 \institution{MIT}}
\email{xiuming@csail.mit.edu}


\author{Ren Ng}
\affiliation{%
 \institution{University of California, Berkeley}}
\email{ren@berkeley.edu}

\author{David E. Jacobs}
\affiliation{%
 \institution{Google Research}}
\email{dejacobs@google.com}

\renewcommand\shortauthors{Zhang, X. et al}

\newcommand{\dummyfig}[1]{
  \centering
  \fbox{
    \begin{minipage}[c][0.33\textheight][c]{0.8\textwidth}
      \centering{#1}
    \end{minipage}
  }
}

\begin{abstract}
Casually-taken portrait photographs often suffer from unflattering lighting and shadowing because of suboptimal conditions in the environment.
Aesthetic qualities such as the position and softness of shadows and the lighting ratio between the bright and dark parts of the face are frequently determined by the constraints of the environment rather than by the photographer.
Professionals address this issue by adding light shaping tools such as scrims, bounce cards, and flashes.
In this paper, we present a computational approach that gives casual photographers some of this control, thereby allowing poorly-lit portraits to be relit post-capture in a realistic and easily-controllable way.
Our approach relies on a pair of neural networks---one to remove foreign shadows cast by external objects, and another to soften facial shadows cast by the features of the subject and to add a synthetic fill light to improve the lighting ratio.
To train our first network we construct a dataset of real-world portraits wherein synthetic foreign shadows are rendered onto the face, and we show that our network learns to remove those unwanted shadows.
To train our second network we use a dataset of Light Stage scans of human subjects to construct input/output pairs of input images harshly lit by a small light source, and variably softened and fill-lit output images of each face. We propose a way to explicitly encode facial symmetry and show that our dataset and training procedure enable the model to generalize to images taken in the wild.
Together, these networks enable the realistic and aesthetically pleasing enhancement of shadows and lights in real-world portrait images.\footnote{\url{https://people.eecs.berkeley.edu/~cecilia77/project-pages/portrait}}
\end{abstract}

%

\ccsdesc[500]{Computing methodologies~Computational photography}

%
%

\keywords{computational-photography}

\begin{teaserfigure}
\begin{center}\centering
    \setlength{\tabcolsep}{0.05cm}
    \setlength{\itemwidth}{3.4cm}
    \begin{tabular}{c@{\hspace{0.1cm}}ccccc}
    \rotatebox{90}{\hspace{14mm}Input}&
      \includegraphics[width=\itemwidth]{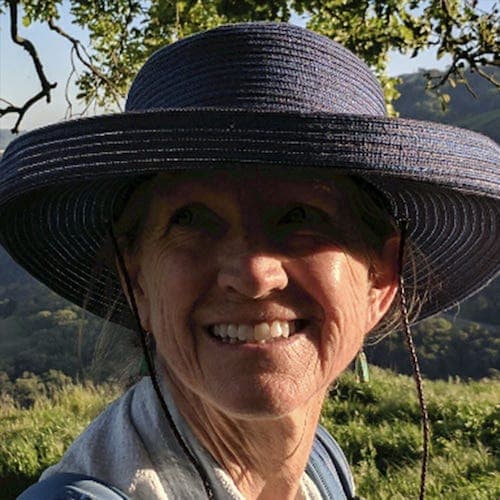}&
       \includegraphics[width=\itemwidth]{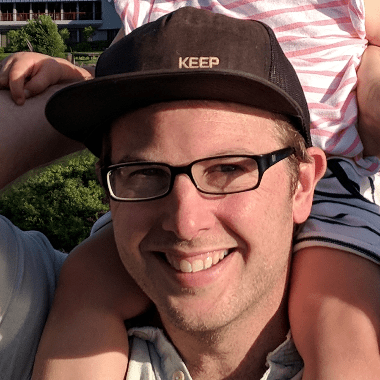}&
      \includegraphics[width=\itemwidth]{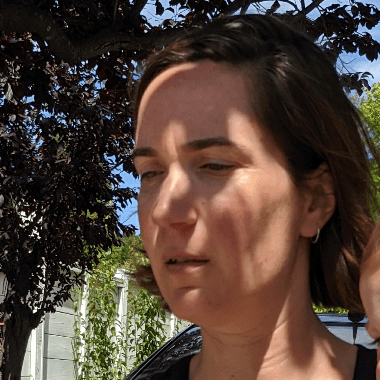}&
      \includegraphics[width=\itemwidth]{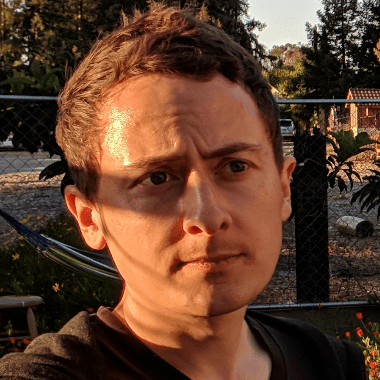}&
      \includegraphics[width=\itemwidth]{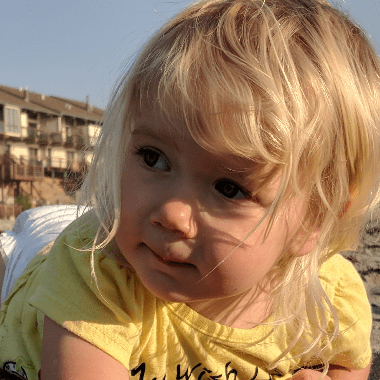}\\
      \rotatebox{90}{\hspace{2mm}Our enhanced output}&
      \includegraphics[width=\itemwidth]{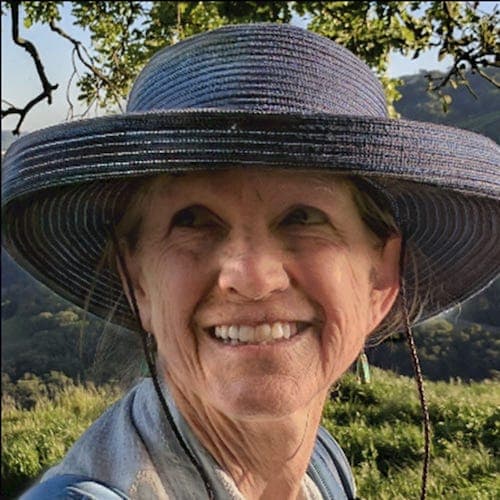}&
      \includegraphics[width=\itemwidth]{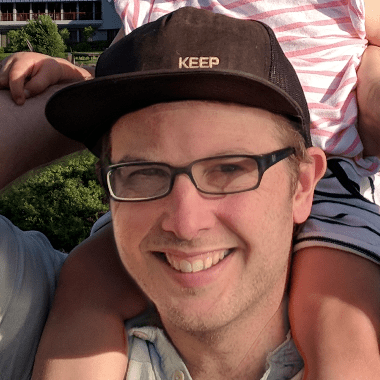}&
      \includegraphics[width=\itemwidth]{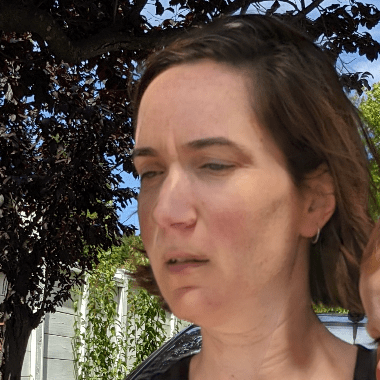}&
      \includegraphics[width=\itemwidth]{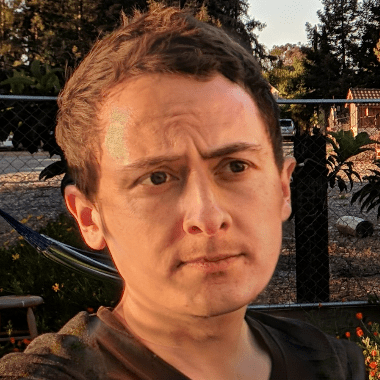}&
      \includegraphics[width=\itemwidth]{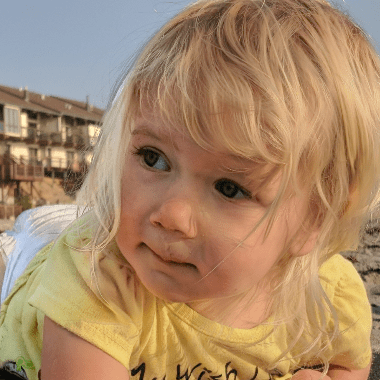}\\
    \end{tabular}\vspace{-0.2cm}
    \vspace{-0.2cm}
	\captionof{figure}{
	\label{fig:teaser}
	The results of our portrait enhancement method on real-world portrait photographs. Casual portrait photographs often suffer from undesirable shadows, particularly foreign shadows cast by external objects, and dark facial shadows cast by the face upon itself under harsh illumination.
	We propose an automated technique for enhancing these poorly-lit portrait photographs by removing unwanted foreign shadows, reducing harsh facial shadows, and adding synthetic fill lights.}
\end{center}
\end{teaserfigure}

\maketitle

\section{Introduction}
\label{sec:intro}

The aesthetic qualities of a photograph are largely influenced by the interplay between light, shadow, and the subject.
By controlling these scene properties, a photographer can alter the mood of an image, direct the viewer's attention, or tell a specific story. Varying the position, size, or intensity of light sources in an environment can affect the perceived texture, albedo, and even three-dimensional shape of the subject. This is especially true in portrait photography, as the human visual system is particularly sensitive to subtle changes in the appearance of human faces.
For example, soft lighting (\eg light from a large area light source like an overcast sky) reduces skin texture, which may cause the subject to appear younger. Conversely, harsh lighting (\eg light from a small or distant source like the midday sun) may exaggerate wrinkles and facial hair, making a subject appear older. Similarly, any shadows falling on the subject's face can accentuate its three-dimensional structure or obfuscate it with distracting intensity edges that are uncorrelated with facial geometry.
Other variables such as the lighting angle (the angle at which light strikes the subject) or the lighting ratio (the ratio of illumination intensity between the brightest and darkest portion of a subject's face) can affect the dramatic quality of the resulting photograph, or may even affect some perceived quality of the subject's personality: harsh lighting may look ``serious'', or lighting from below may make the subject look ``sinister''.

Unfortunately, though illumination is clearly critical to the appearance of a photograph, finding or creating a good lighting environment outside of a studio is challenging.
Professional photographers spend significant amounts of time and effort directly modifying the illumination of existing environments through physical means, such as elaborate lighting kits consisting of scrims (cloth diffusers), reflectors, flashes, and bounce cards~\cite{grey2014master}.

In this work, we attempt to provide some of the control over lighting that professional photographers have in studio environments to casual photographers in unconstrained environments.
We present a framework that allows casual photographers to enhance the quality of light and shadow in portraits from a single image after it has been captured. We target three specific lighting problems common in casual photography and uncontrolled environments:
\paragraph{Foreign shadows}
We will refer to any shadow cast on the subject's face by an external occluder (\eg a tree, a hat brim, an adjacent subject in a group shot, the camera itself, etc.) as a \textit{foreign shadow}. Notably, foreign shadows can result in an arbitrary two-dimensional shape in the final photograph, depending on the shape of the occluder and position of the primary, or \textit{key}, light source. Accordingly, they frequently introduce image intensity edges that are uncorrelated with facial geometry and therefore are almost always distracting.
Because most professional photographers would remove the occluder or move the subject in these scenarios, we will address this type of shadow by attempting to remove it entirely.
\paragraph{Facial shadows}
We will refer to any shadow cast on the face by the face itself (\eg the shadow attached to the nose when lit from the side) as a \textit{facial shadow}.
    Because facial shadows are generated by the geometry of the subject, 
    these shadows (unlike foreign shadows) can only project to a small space of two-dimensional shapes in the final image.
    Though they may be aesthetically displeasing, the image intensity edges introduced by facial shadows are more likely than foreign shadows to be a meaningful cue for the shape of the subject.
    Because facial shadows are almost always present in natural lighting environments (\ie the environment is not perfectly uniform), we do not attempt to remove them entirely.
    We instead emulate a photographer's scrim in this scenario, which effectively increases the size of the key light and softens the edges of the shadows it casts.
\paragraph{Lighting ratios}
    In scenes with very strong key lights (\eg outdoors on a clear day), the ratio between the illumination of the brightest and darkest parts of the face may exceed the dynamic range of the camera, resulting in a portrait with dark shadows or blown out highlights.
    While this can be an intentional artistic choice, typical portrait compositions target less extreme lighting ratios.
    Professional photographers balance lighting ratios by placing a secondary, or \textit{fill}, light in the scene opposite the key.
    We similarly place a virtual fill light to balance the lighting ratio and add definition to the shape of the shadowed portion of the subject's face.

Our framework consists of two machine learning models: one trained for foreign shadow removal, and another trained for handling facial shadow softening and lighting ratio adjustment. This grouping of tasks is motivated by two related observations.

Our first observation, as mentioned above, is tied to the differing relationships between shadow appearance and facial geometry.
The appearance of facial shadows in the input image provides a significant cue for shape estimation, and should therefore be useful input when synthesizing an image with softer facial shadowing and a smaller lighting ratio.
But foreign shadows are much less informative, and so we first identify and remove all foreign shadows before attempting to perform facial shadow manipulation.
This approach provides our facial shadow model with an image in which all shadow-like image content is due to facial shadows, and also happens to be consistent with contemporary theories on how the human visual system perceives shadows~\cite{rensink2004influence}.

Our second observation relates to training dataset requirements.

Thanks to the unconstrained nature of foreign shadow appearance, it is possible to train our first network with a synthetic dataset: 5000 ``in-the-wild'' images, augmented with randomly generated foreign shadows for a total of 500K training examples.
This strategy is not viable for our second network, as facial shadows must be consistent with the geometry of the subject and so cannot be generated in this way. Constructing an ``in-the-wild'' dataset consisting entirely of images with controlled facial shadowing is also intractable.
We therefore synthesize the training data for this task using one-light-at-a-time (OLAT) scans taken by a Light Stage, an acquisition setup and method proposed to capture reflectance field~\cite{debevec2000acquiring} of human faces. We use the Light Stage scans to synthesize paired harsh/soft images for use as training data.
Section~\ref{sec:data} will discuss our dataset generation procedure in more detail.

Though trained separately, the neural networks used for our two tasks share similar architectures: both are deep convolutional networks for which the input is a $256 \times 256$ resolution RGB image of a subject's face.
The output of each network is a per-pixel and per-channel affine transformation consisting of a scaling $A$ and offset $B$, at the same resolution as the input image $\myimage_{in}$ such that the final output $\myimage_{out}$ can be computed as:
\begin{equation}
    \myimage_{out} = \myimage_{in} \circ A + B,
    \label{eq:affine}
\end{equation}
where $\circ$ denotes per-element multiplication. This approach can be thought of as an extension of quotient images~\cite{shashua2001quotient} and of residual skip connections~\cite{he2016deep}, wherein our network is encouraged to produce output images that resemble scaled and shifted versions of the input image.
The facial shadow network includes additional inputs that are concatenated onto the input RGB image: 1) two numbers encoding the desired shadow softening and fill light brightness, so that variable amounts of softening and fill lighting can be specified and 2) an additional rendition of the input image with the face mirrored about its axis of symmetry (\ie pixels corresponding to the left eye of the input are warped to the position of the right eye, and vice versa).
Using a mirrored face image in this way broadens the spatial support of the first layer of our network to include the image region on the opposite side of the subject's face.
This allows the network to exploit the bilateral symmetry of human faces and to easily ``borrow'' pixels with similar semantic meaning and texture but different lighting from the opposite side of the subject's face (see  Section~\ref{sec:facial-sym} for details).

In addition to the framework itself, this work presents the following technical contributions:
\begin{compactenum}
    \item Techniques for generating synthetic, real-world, and Light Stage-based datasets for training and evaluating machine learning models targeting foreign shadows, facial shadows, and virtual fill lights.
    \item Symmetric face image generation for explicitly encoding symmetry cue into training our facial shadow model.
    \item Ablation studies that demonstrate our data and models achieve portrait enhancement results that outperform all baseline methods in numerical metrics and perceptual quality.
\end{compactenum}

The remainder of the paper is organized as follows. Section~\ref{sec:related} describes prior work in lighting manipulation, shadow removal, and portrait retouching. Section~\ref{sec:data} introduces our synthetic dataset generation procedure and our real ground-truth data acquisition. 
Section~\ref{sec:model} talks about our network architecture and training procedure. Section~\ref{sec:experiments} shows a series of ablation studies and presents qualitative and quantitative results and comparisons. Section~\ref{sec:limitations} discusses limitations of our approach.
\vspace{-0.03cm}
\section{Related Work}
\label{sec:related}

The detection and removal of shadows in images is a central problem within computer vision, as is the closely related problem of separating image content into reflectance and shading~\cite{Horn1974DeterminingLF}.
Many graphics-oriented shadow removal solutions rely on manually-labeled ``shadowed'' or ``lit'' regions~\cite{wu2007natural, shor2008shadow, Gryka2015softShadows, arbel2010shadow}.
Once manually identified, shadows can be removed by solving a global optimization technique, such as graph cuts.
Because relying on user input limits the applicability of these techniques, fully-automatic shadow detection and manipulation algorithms have also attracted substantial attention.
Illumination discontinuity across shadow edges~\cite{sato2003illumination} can be used to detect and remove shadows~\cite{baba2003shadow}. Formulating shadow enhancement as local tone adjustment and using edge-preserving histogram manipulation~\cite{kaufman2012content} enables contrast enhancement on semantically segmented photographs.
Relative differences in the material and illumination of paired image segments~\cite{guo2012paired, ma2016appearance} enables the training of region-based classifiers and the use of graph cuts for labeling and shadow removal.
Shadow removal has also been formulated as an entropy minimization problem~\cite{finlayson2002removing, finlayson2009entropy}, where invariant chromaticity and intensity images are used to produce a shadow mask that is then re-integrated to form a shadow-free image.
These methods assume that shadow regions contain approximately constant reflectance and that image gradients are entirely due to changes in illumination, and are thereby fail when presented with complex spatially-varying textures or soft shadowing.
In addition, by decomposing the shadow removal problem into two separate stages of detection and manipulation, these methods cannot recover from errors during the shadow detection step~\cite{ma2016appearance}.

General techniques for inverse rendering~\cite{ramamoorthi2001signal,sfsnetSengupta18} and intrinsic image decomposition~\cite{grosse2009ground,BarronTPAMI2015} should, in theory, be useful for shadow removal, as they provide shading and reflectance decompositions of the image. However, in practice these techniques perform poorly when used for shadow removal (as opposed to shading removal) and usually consider cast shadows to be out of scope.
For example, the canonical Retinex algorithm~\cite{Horn1974DeterminingLF} assumes that shading variation is smooth and monochromatic and therefore fails catastrophically on simple cases such as shadows cast by the midday sun, which are usually non-smooth and chromatic (sunlit yellow outside the shadow, and sky blue within).

More recently, learning-based approaches have demonstrated a significant improvement on general-purpose shadow detection and manipulation~\cite{khan2015automatic,hu2018direction, hu2019direction,ding2019argan,zhu2018bidirectional,zheng2019distraction,cun2019ghostfree}.
However, like all learned techniques, such approaches are limited by the nature of their training data. While real-world datasets for general shadow removal are available~\cite{Qu_2017_CVPR,Wang_2018_CVPR}, they do not include human subjects and therefore are unlikely to be useful for our task, which requires the network to reason about specific visual characteristics of faces, such as the skin's subsurface scattering effect~\cite{donner2006spectral}. 
Instead, in this paper, we propose to train a model using synthetic shadows generated on images in the wild. We only use images of faces to encourage the model to learn and use priors on human faces. Earlier work has shown that training models on faces improves performance on face-specific subproblems of common tasks, such as inpainting~\cite{yeh2017semantic,ulyanov2018deep}, super-resolution~\cite{chen2018fsrnet} and synthesis~\cite{denton2015deep}.

Another problem related to ours is ``portrait relighting''---the task of relighting a single image of a human subject according to some desired environment map~\cite{Sun2019, zhou2019deep}.
These techniques could theoretically be used for our task, as 
manipulating the facial shadows of a subject is equivalent to re-rendering that subject under a modified environmental illumination map in which the key light has been dilated.
However, as we will demonstrate (and was noted in~\cite{Sun2019}) these techniques struggle when presented with images that contain foreign shadows or high-frequency image structure due to harsh shadows in the input image, which our approach specifically addresses. Example-based portrait lighting transfer techniques~\cite{Shih2014, shu2017portrait} also represent potential alternative solutions to this task, but they require a high-quality reference image that exhibits the desired lighting, and that also contains a subject with a similar identity and pose as the input image---an approach that  does not scale to casual photos in the wild. 

\begin{figure*}[ht!]
\includegraphics[width=\linewidth]{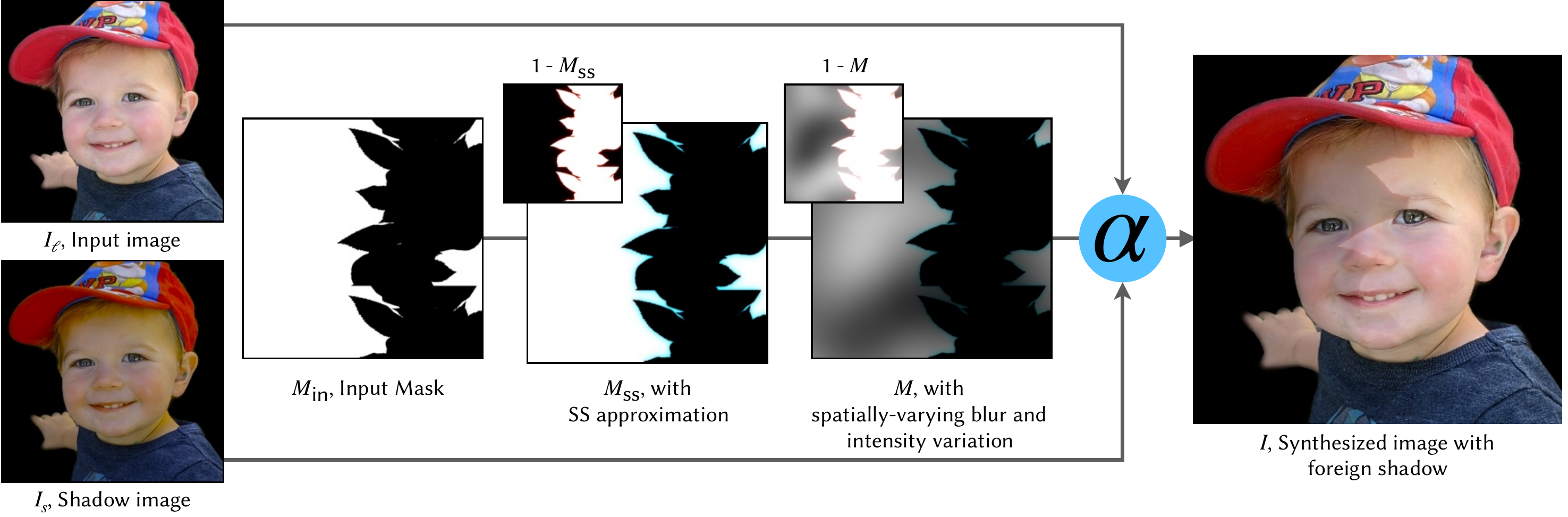}
\vspace{-0.2cm}
\captionof{figure}{
The pipeline of our foreign shadow synthesis model (Section~\ref{sec:foreign-syn}). The colors of the ``lit'' image $\myimage_{\light}$ are randomly jittered to generate a ``shadow'' image $\myimage_s$. The input mask $M_{\texttt{in}}$ shown here is generated from an object silhouette, though it may also be generated with Perlin noise. $M_{\texttt{in}}$ is subjected to a subsurface scattering (SS) approximation of human skin to generate $M_{\texttt{ss}}$, then a spatially-varying blur and per-pixel intensity variation to generate $M$. $\myimage_{\light}$ and $\myimage_s$ are then blended according to the shadow mask $M$ to generate a training sample $\myimage$.}
\label{fig:foreign-syn}
\end{figure*}

\section{Data Synthesis}
\label{sec:data}
There is no tractable data acquisition method to collect a large-scale dataset of human faces for our task with diversity in both the subject and the shadows, as the capture process would be onerous for both the subjects (who must remain perfectly still for impractically long periods of time) and the photographers (who must be specially trained for the task and find thousands of willing participants in thousands of unique environments).
Instead, we synthesize custom datasets for our subproblems by augmenting existing datasets---Recall that our two models require fundamentally different training data. Our foreign shadow datasets (Section~\ref{sec:foreign-syn}) are based on images of faces in the wild with rendered shadows, while our facial shadow and fill light datasets (Section~\ref{sec:facial-syn}) are based on a Light Stage dataset with carefully chosen simulated environments.

\subsection{Foreign Shadows}
\label{sec:foreign-syn}
To synthesize images that appear to contain foreign shadows, we model images as a linear blend between a ``lit'' image $\myimage_{\light}$ and a ``shadowed'' image $\myimage_{s}$, according to some shadow mask $M$:
\begin{equation}
    \myimage = \myimage_{\light} \circ (1 - M) + \myimage_{s} \circ M
    \label{eq:foreign}
\end{equation}
The lit image $\myimage_{\light}$ is assumed to contain the subject lit by all light sources in the scene (\eg the sun and the sky), and the shadowed image $\myimage_{s}$ is assumed to be the subject lit by everything other than the key (\eg just the sky). The shadow mask $M$ indicates which pixels are shadowed: $M=1$ if fully shadowed, and $M=0$ if fully lit. To generate a training sample, we need $\myimage_{\light}$, $\myimage_{s}$, and $M$. $\myimage_{\light}$ is selected from an initial dataset described below, $\myimage_{s}$ is a color transform of $\myimage_{\light}$, and $M$ comes from a silhouette dataset or a pseudorandom noise function.

Because deep learning models are highly sensitive to the realism and biases of the data used during training, we take great care to synthesize as accurate a shadow mask and shadowed image as possible with a series of augmentations on $\myimage_{s}$ and $M$. Figure~\ref{fig:foreign-syn} presents an overview of the process and below we enumerate different aspects of our synthesis model and their motivation. In Section~\ref{sec:exp-ablation}, we will demonstrate their efficacy through an ablation study.
\paragraph{Input Images}
Our dataset is based on a set of 5,000 faces in the wild that we manually identified as not containing any foreign shadows. These images are real, in-the-wild JPEG data, and so they are varied in terms of subject, ethnicity, pose, and environment. Common accessories such as hats and scarves are included, but only if they do not cast shadows. We make one notable exception to this policy: glasses. Shadows from glasses are unavoidable and behave more like facial shadows than foreign. Accordingly, shadows from glasses are preserved in our ground truth. Please refer to the supplement for examples.
\paragraph{Light Color Variation}
The shadowed image region $\myimage_{s}$ is illuminated by a lighting environment different from that of the non-shadow region. For example, outdoor shadows are often tinted blue because when the sun is blocked, the blue sky becomes the dominant light source. To account for such illumination differences, we apply a random color jitter, formulated as a $3 \times 3$ color correction matrix, to the lit image $\myimage_{\light}$. Please see the supplement for details.
\paragraph{Shape Variation}
The shapes of natural shadows are as varied as the shapes of natural objects in the world, but those natural shapes also exhibit significant statistical regularities~\cite{huang2000statistics}. To capture both the variety and the regularity of real-world shadows, our distribution of input shadow masks $M_{\texttt{in}}$ is half ``regular'' real-world masks drawn from a dataset of 300 silhouette images of natural objects, randomly scaled and tiled; and half ``irregular'' masks generated using a Perlin noise function at 4 octaves with a persistence drawn uniformly at random within $[0, 0.85]$, with the initial amplitude set to 1.0.
\begin{figure*}
\includegraphics[width=\linewidth]{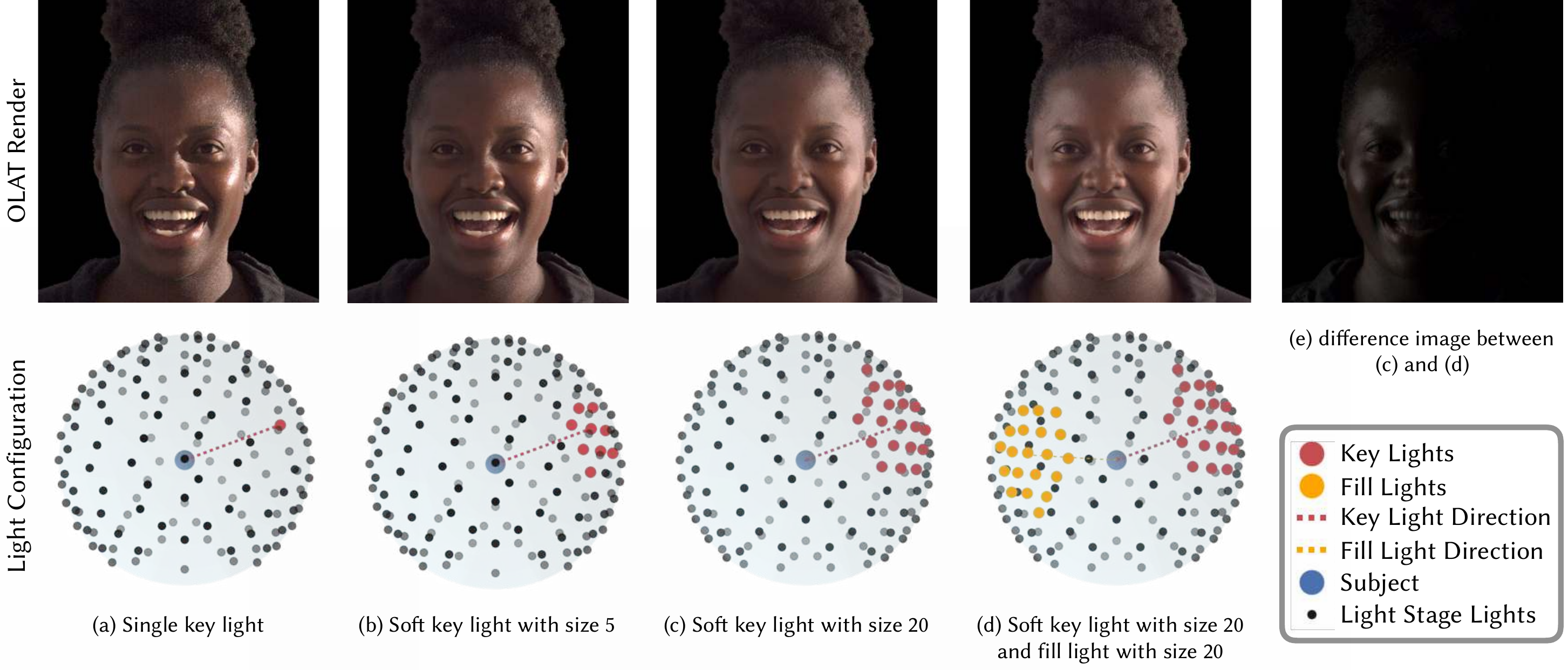}
\captionof{figure}{Our facial shadow synthesis model. Our input image is a OLAT render corresponding to an environment with a single key light turned on as shown in (a). To soften the shadows by some variable amount, we distribute the key's energy to a variable number of its neighbors, as shown in (b) and (c). We also add a number of fill lights on the opposite side of the Light Stage, to brighten the darker side of the face as shown in (d), with the fill light's difference image visualized in (e). For clarity, only half of the Light Stage's lights are rendered. See the supplemental video at 02:04 for a few more examples.}
\label{fig:facial-syn}
\end{figure*}
\paragraph{Subsurface Scattering}
Light scatters beneath the surface of human skin before exiting, and the degree of that scattering is wavelength-dependent~\cite{hanrahan1993reflection, jensen2001practical,krishnaswamy2004biophysically}: blood vessels cause red light to scatter further that other wavelengths, causing a visible color fringe at shadows. We approximate the subsurface scattering appearance by uniformly blurring $M_{\texttt{in}}$ with a different kernel per color channel, borrowing from \citet{fernando2004gpu}. In brief, the kernel for each channel is a sum of Gaussians $G(\sigma_{c, k})$ with weights $w_{c, k}$, such that each channel $M_{c}$ of the shadow mask with subsurface scattering $M_{\texttt{ss}}$ is rendered as:
\begin{equation}
M_{c} = \sum_{k} M_{\texttt{in}} * G(\sigma_{c, k}) w_{c, k}.
\end{equation}
\paragraph{Spatial Variation}
The softness of the shadow being cast on a subject depends on the relative distances between the subject, the key light, and the object casting the shadow. Because this relationship varies over the image, our shadow masks incorporate a spatially-varying blur over $M_{\texttt{ss}}$.
While many prior works assume that the shadow region has a constant intensity~\cite{zhang2019effective}, we note that a partially translucent occluder or an environment violating the assumption that lights are infinitely far away will cause shadows to have different local intensities. Accordingly, we similarly apply a spatially-varying per-pixel intensity variation to $M_{\texttt{ss}}$ as well, modeled as Perlin noise at 2 octaves with a persistence drawn uniformly at random from $[0.05, 0.25]$ and an initial amplitude set to 1.0. The final mask with spatial variation incorporated is what we refer to as $M$ above.

\subsection{Facial Shadows}
\label{sec:facial-syn}
We are not able to use ``in-the-wild'' images for synthesizing \emph{facial} shadows because the highly accurate facial geometry it would require is generally not captured in such datasets.
Instead, we use Light Stage data that can relight a scanned subject with perfect fidelity under any environment and select the simulated environment with care.
Note that we \emph{cannot} use light stage data to produce more accurate foreign shadows than we could using raw, in-the-wild JPEG images, which is why we adopt different data synthesis for these two tasks.

When considering \emph{foreign} shadows, we adopt shadow \emph{removal} with the rationale that foreign shadows are likely undesirable from a photographic perspective and removing them does not affect the apparent authenticity of the photograph (as the occluder is rarely in frame). \emph{Facial} shadows, in contrast, can only be \emph{softened} if we wish to affect the mood of the photograph while remaining faithful to the scene's true lighting direction. 

We construct our dataset by emulating the scrims and bounce cards employed by professional photographers. Specifically, we generate harsh/soft facial shadow pairs using OLAT scans from a Light Stage dataset. This is ideal for two reasons: 1) each individual light in the stage is designed to match the angular extent of the sun, so it is capable of generating harsh shadows, and 2) with such a dataset, we can render an image $\myimage$ simulating an arbitrary lighting environment with a simple linear combination of OLAT images $\myimage_i$ with weights $w_i$, \ie $\myimage = \sum_i \myimage_i w_i$.

For each training instance, we select one of the $304$ lights in the stage and dub it our key light with index $i_{\texttt{key}}$, and use its location to define the key light direction $\myvec{\light}_{\texttt{key}}$.
Our harsh input image is defined to be one corresponding to OLAT weights
${w_i = \{\keyintensity \text{ if } i = i_{\texttt{key}}}$, ${\epsilon \text{ otherwise}\}}$,
where $\keyintensity$ is a randomly sampled intensity of the key light and $\epsilon$ is a small non-zero value that adds ambient light to prevent shadowed pixels from becoming fully black.
The corresponding soft image is then rendered by splatting the key light energy to the set of its $\lightcount$ nearest neighboring lights $\neighborhood(\myvec{\light}_{\texttt{key}})$, where $\lightcount$ is drawn uniformly from a set of discrete numbers $[5, 10, 20, 30, 40]$. This can be thought of as convolving the key light source with a disc, similar in spirit to a diffuser or softbox. We then compute the location of the fill light (Figure~\ref{fig:facial-syn}(d)):
\begin{equation}
\myvec{\light}_{\texttt{fill}} = 2 (\myvec{\light}_{\texttt{key}} \cdot \myvec{n}) \myvec{n} - \myvec{\light}_{\texttt{key}},
\end{equation}
where $\myvec{n}$ is the unit vector along the camera $z$-axis, pointing out of the Light Stage. For all data generation, we use a fixed fill light neighborhood size of 20, and a random fill intensity $\fillintensity$ in $[0, \keyintensity/10]$. Thus, the soft output image is defined as one corresponding to OLAT weights
\begin{equation*}
w_i = \left\{\begin{array}{rl}
         \keyintensity/\lightcount, & \text{if } i \in \neighborhood(\myvec{\light}_{\texttt{key}})\\
         \fillintensity, & \text{if } i \in \neighborhood(\myvec{\light}_{\texttt{fill}})\\
         \epsilon, & \text{otherwise}
    \end{array}
\right.
\end{equation*}

To train our facial shadow model, we use OLAT images of 85 subjects, each of which was imaged under different expressions and poses, giving us in total 1795 OLAT scans to render our facial harsh shadow dataset. We remove degenerate lights that cause strong flares or at extreme angles that render too dark images, and end up using the remaining 284 lights for each view.

\subsection{Facial Symmetry}
\label{sec:facial-sym}

\newcommand{\distmat}{\mathrm{D}}
\newcommand{\weightmat}{\mathrm{W}}
\newcommand{\pointx}{x}
\newcommand{\pointy}{y}
\newcommand{\vertx}{u}
\newcommand{\verty}{v}

\begin{figure}
        \setlength{\itemwidth}{3.8cm}
        \begin{tabular}{c@{\quad\,}c}
          \includegraphics[width=\itemwidth]{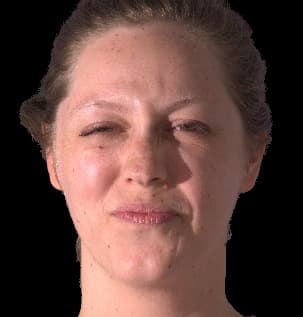}&
          \includegraphics[width=\itemwidth]{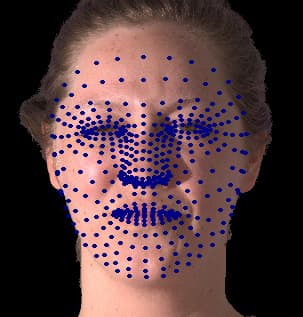} \\
            \small (a) Input Image &
            \small (b) Detected Landmarks \\
          \includegraphics[width=\itemwidth]{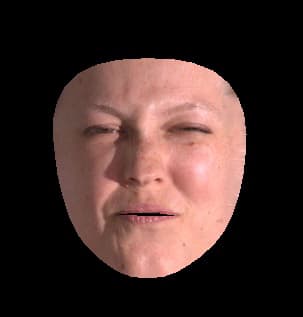}&
          \includegraphics[width=\itemwidth]{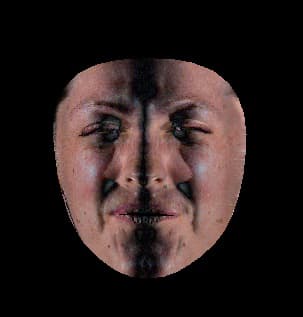} \\
            \small (c) Mirrored Input &
            \small (d) |Input - Mirror| \\
        \end{tabular}\vspace{-0.2cm}
\captionof{figure}{
The symmetry of human faces is a useful cue for reasoning about lighting: a face's reflectance and geometry is likely symmetric, but the shadow cast upon that face is likely not symmetric. To leverage this, a landmark detection system is applied to the input image (a) and the recovered landmark (b) are used to produce a per-pixel mirrored version of the input image (c). This mirrored image is appended to the input image in our networks, which improves performance by allowing the network to directly reason about asymmetric image content (d) which is likely due to facial and foreign shadows.}

\label{fig:facial-sym}
\vspace{-0.3cm}
\end{figure}

Human faces tend to be bilaterally symmetric: the left side of most faces closely resembles the right side in terms of geometry and reflectance, except for the occasional blemish or minor asymmetry. However, images of faces are rarely symmetric because of facial shadows. Therefore, if a neural network can easily reason about the symmetry of image content on the subject's face, it will be able to do a better job of reducing shadows cast upon that face. For this reason, we augment the image that is input to our neural networks with a ``mirrored'' version of that face, thereby giving the early layers of those networks the ability to straightforwardly reason about which image content is present on the opposite side of the face. Because the subject's face is rarely perfectly vertical and oriented perpendicularly to the camera's viewing direction, it is not sufficient to simply mirror the input image along the $x$-axis. We therefore estimate the geometry of the face and mirror the image using that estimated geometry, by warping image content near each vertex of a mesh to the location of its corresponding mirrored vertex. See Figure~\ref{fig:facial-sym} for a visualization. 

Given an image $\myimage$, we use the landmark detection system of \cite{kartynnik2019real} to produce a model of facial geometry consisting of 468 2D vertices (Figure~\ref{fig:facial-sym}(b)) and a mesh topology (which is fixed for all instances). For each vertex $j$ we precompute the index of its bilaterally symmetric vertex $\bar{j}$, which corresponds to a vertex $(\vertx_{\bar{j}}, \verty_{\bar{j}})$ at the same position as $(\vertx_j, \verty_j)$ but on the opposite side of the face. With this correspondence we could simply produce a ``mirrored'' version of $I$ by applying a meshwarp to $I$ where the position of each vertex $j$ is moved to the position of its mirror vertex $\bar j$. However, a straightforward meshwarp is prone to triangular-shaped artifacts and irregular behavior on foreshortened triangles or inaccurately-estimated keypoint locations. For this reason we instead use a ``soft'' warping approach based on an adaptive radial basis function (RBF) kernel: For each pixel in $I$ we compute its RBF weight with respect to the 2D locations of all vertices, express that pixel location as a convex combination of all vertex locations, and then interpolate the ``mirrored'' pixel location by computing the same convex combination of all \emph{mirrored} vertex locations. Put formally, we first compute the Euclidean distance from all pixel locations to all vertex locations:
\begin{equation}
    \distmat_{i,j} = \left( \pointx_i - \vertx_j \right)^2 + \left( \pointy_i - \verty_j \right)^2
\end{equation}
With this we compute a weight matrix consistent of normalized Gaussian distances:
\begin{equation}
\weightmat_{i,j} = \frac{\exp\left(-\distmat_{i,j} / \sigma_j \right)}{\sum_{j'}\exp\left(-\distmat_{i,j'} / \sigma_{j'} \right)}
\end{equation}
Unlike a conventional normalized RBF kernel, $\weightmat_{i,j}$ is computed using a different $\sigma$ for each of the $j$ vertices. Each vertex's $\sigma$ is selected such that each landmark's influence in the kernel is inversely proportional to how many nearby neighbors it has for this particular image:
\begin{equation}
\sigma_j = \select_{j'}\left(\left( \vertx_j - \vertx_{j'} \right)^2 + \left( \verty_j - \verty_{j'} \right)^2, K_\sigma\right)
\end{equation}
Where $\select(\cdot, K)$ returns the $K$'th smallest element of an input vector. This results in a warp where sparse keypoints have significant influence over their local neighborhood, while the influence of densely packed keypoints is diluted. This weight matrix is then used to compute the weighted average of mirrored vertex locations, and this 2D location is used to bilinearly interpolate into the input image to produce it's mirrored equivalent:
\begin{equation}
\bar{\myimage} = \myimage \left( \sum_j \weightmat_{i,j} \vertx_{\bar{j}}\,,\, \sum_j \weightmat_{i,j} \verty_{\bar{j}}
\right)
\end{equation}
The only hyperparameter in this warping model is an integer value $K_\sigma$, which we set to $4$ in all experiments. This proposed warping model is robust to asymmetric expressions and poses assuming the landmarks are accurate, but is sensitive to asymmetric skin features, e.g., birthmarks.

The input to our facial shadow network is the concatenation of the input image $\myimage$ with its mirrored version $\bar{\myimage}$ along the channel dimension. This means that the receptive field of our CNN includes not just the local image neighborhood, but also its mirrored counterpart. Note that we do not include the mirrored image as input to our foreign shadow model, as we found it did not improve results. We suspect that this is due to the unconstrained nature of foreign shadow appearance, which weakens the assumption that corresponding face regions will have different lighting.

\section{Neural Network Architecture and Training}
\label{sec:model}
Here we describe the neural network architectures that we use for removing foreign shadows and for softening facial shadows. As the two tasks use different datasets and there is an additional conditional component in the facial shadow softening model, we train these two tasks separately.

For both models, we employ a GridNet~\cite{fourure2017residual} architecture with modifications proposed in~\cite{niklaus2018context}. GridNet is a grid-like architecture of rows and columns, where each row is a stream that processes features with resolution kept unchanged, and columns connect the streams by downsampling or upsampling the features. By allowing computation to happen at different layers and different spatial scales instead of conflating layers and spatial scales (as U-Nets do) GridNet produces more accurate predictions as has been successfully applied to a number of image synthesis tasks~\cite{niklaus2018context,niklaus20193d}. We use a GridNet with eight columns wherein the first three columns perform downsampling and the remaining five columns perform upsampling, and use five rows for foreign model and six rows for facial model, as we found this to work best after an architecture search. 

For all training samples, we run a face detector to obtain a face bounding box, then resize and crop the face into $256\times256$ resolution. For the foreign shadow removal model, the input to the network is a 3-channel RGB image and the output of the model is a $3$-channel scaling $A$ and a $3$-channel offset $B$, which are then applied to the input to produce a 3-channel output image (Equation~\ref{eq:affine}). For the facial shadow softening model, we additionally concatenate the input to the network with its mirrored counterpart (as per Section~\ref{sec:facial-sym}). As we would like our model to allow for a \emph{variable} degree of shadow softening and of fill lighting intensity, we introduce two ``knobs''---one for light size $\lightcount$ and the other for fill light intensity $\fillintensity$, which are assumed to be provided as input. To inject this information into our network, a 2-channel image containing these two values at every pixel is concatenated into both the input and the last layers of the encoders of the network.

We supervise our two models using a weighted combination of pixel-space L1 loss ($\lossfun{pix}$) and a perceptual feature space loss ($\lossfun{feat}$) which has been used successfully to train models such as image synthesis and image decomposition~\cite{chen2017photographic,xzhang2018reflectionremoval,zhang2019synthetic}.
Intuitively, the perceptual loss accounts for high-level semantics in the reconstructed image but may be invariant to some non-semantic image content. By additionally minimizing a per-pixel L1 loss our model is better able to recover low-frequency image content.
The perceptual loss is computed by processing the reconstructed and ground truth images through a pre-trained VGG-19 network $\Phi(\cdot)$ and computing the L1 difference between extracted features in selected layers as specified in~\cite{xzhang2018reflectionremoval}.
The final loss function is formulated as:
\begin{align}
    \lossfun{feat}(\theta) &= \sum_d \lambda_d \norm{\Phi_d \left(\myimage^* \right) - \Phi_d\left(f\left(\myimage_\mathit{in}; \theta\right)\right)}_1  \nonumber  \\
    \lossfun{pix}(\theta) &= \norm{\myimage^* - f(\myimage_\mathit{in}; \theta)}_1  \nonumber  \\
    \lossfun{}(\theta) &= 0.01 \times \lossfun{feat}(\theta) + \lossfun{pix}(\theta),
\end{align}
where $\myimage^*$ is the ground-truth shadow-removed or shadow-softened RGB image, $f(\cdot;\theta)$ denotes our neural network, and $\lambda_d$ denotes the selected weight for the $d$-th VGG layer. $\myimage_\mathit{in} = \myimage$ for foreign removal model and $\myimage_\mathit{in} = \operatorname{concat}(\myimage, \bar{\myimage}, \fillintensity, \lightcount)$ for facial shadow softening model.
This same loss is used to train both models separately. We minimize $\lossfun{}$ with respect to both of our model weights $\theta$ using Adam~\cite{KingmaB15} ($\beta_1=0.9$, $\beta_2=0.999$, $\epsilon=10^{-8}$) for 500K iterations, with a learning rate of $10^{-4}$ that is decayed by a factor of $0.9$ every 50K iterations.

\section{Experiments}
\label{sec:experiments}
We use synthetic and real in-the-wild test sets to evaluate our foreign shadow removal model (Section~\ref{sec:exp-foreign-eval}) and our facial shadow softening model (Section~\ref{sec:exp-facial-eval}). We also present an ablation study of the components of our foreign shadow synthesis model  (Section~\ref{sec:exp-ablation}) as well as of our facial symmetry modeling. Extensive additional results can be found in the supplement.

\subsection{Evaluation Data}
\label{sec:exp-eval-data}

\begin{figure}
    \begin{center}
        \setlength{\tabcolsep}{0.05cm}
        \setlength{\itemwidth}{2.7cm}
        \begin{tabular}{ccc}
          \includegraphics[width=\itemwidth]{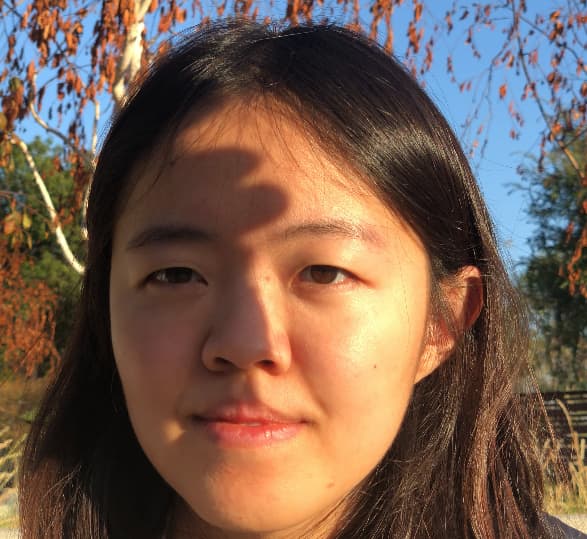}&
          \includegraphics[width=\itemwidth]{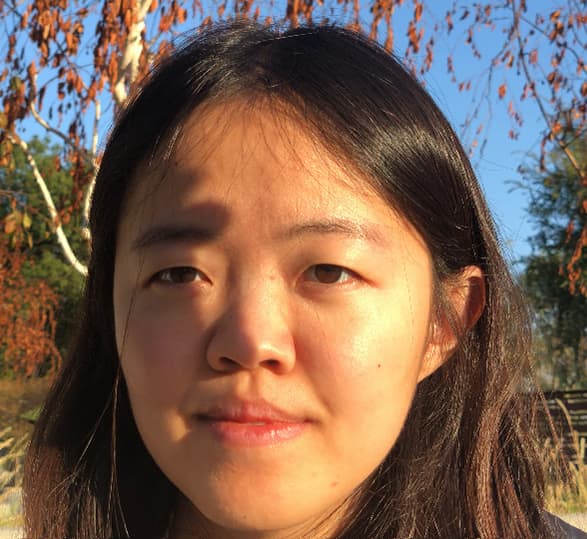}&
          \includegraphics[width=\itemwidth]{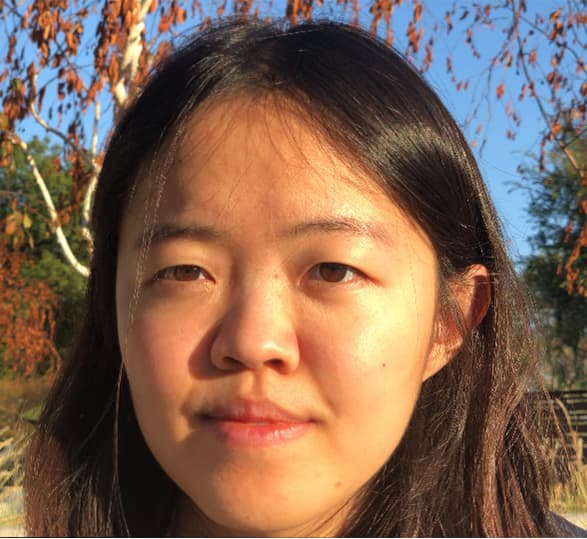}
          \\
          \footnotesize (a) Frame 33 (Shadow 1) &
          \footnotesize (b) Frame 36 (Shadow 2) &
          \footnotesize (c) Frame 54 (Shadow-free)
          \\
        \end{tabular}\vspace{-0.2cm}
    	\captionof{figure}{An example of the shadow removal evaluation dataset we produce using video captured by a smartphone camera stabilized on a tripod.
    	By filming a stationary subject under the shadow cast by a moving foreign occluder (a-c), we are able to obtain multiple input/ground-truth image pairs of the subject (a, c), (b, c).
    	This provides us with an efficient way to collect a set of diverse foreign shadows for evaluation. Please see the supplemental video at 02:39 for more example video clips.}
    	\label{fig:foreign-eval}
    	\vspace{-0.3cm}
    \end{center}
\end{figure}

We evaluate our foreign shadow removal model with two datasets:
\paragraph{foreign-syn}
We use a held-out set of the same synthetic data generation approach described in (Section~\ref{sec:foreign-syn}), where the images (\ie subjects) and shadow masks to generate test-set images are not present in the training set.
\paragraph{foreign-real}
We collect an additional dataset of in-the-wild images for which we can obtain ground-truth images that do not contain foreign shadows. This dataset enables the quantitative and qualitative comparison of our proposed model against prior work. This is accomplished by capturing high-framerate ($60$ fps) videos of stationary human subjects while moving a shadow-casting object in front of the subject. We collect this evaluation dataset outdoors, and use the sun as the dominant light source. For each captured video, we manually identify a set of ``lit'' images and a set of ``shadowed'' images. For each ``shadowed'' image, we automatically use homography to align it to each ``lit'' and find the one that gives the minimum mean pixel error as its counterpart. Because the foreign object is moving during capture, this collection method provides a diversity in the shape and the position of foreign shadows (see examples in Figure~\ref{fig:foreign-eval} and the supplemental video). In total, we capture $20$ videos of $8$ subjects during different times of day, which gives us $100$ image pairs of foreign shadow with ground truth.

We evaluate our facial shadow model with another dataset:
\paragraph{facial-syn}
We use the same OLAT Light Stage data that is used to generate our facial model training data to generate a test set, by using a held-out set of $5$ subjects that are not used during training. We record each harsh input shadow image and the soft ground-truth output image along with their corresponding light size $\lightcount$ and fill light intensity $\fillintensity$ for use. Note that though this dataset is produced through algorithmic means, the ground-truth outputs are a weighted combination of real observed Light Stage images, and are therefore an accurate reflection of the true appearance of the subject up to the sampling limitations of the Light Stage hardware.

\begin{figure*}
\setlength{\tabcolsep}{0.1cm}
\setlength{\itemwidth}{3.35cm}
\begin{tabular}{ccccc}
  \includegraphics[width=\itemwidth]{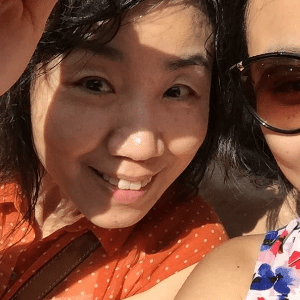}&
  \includegraphics[width=\itemwidth]{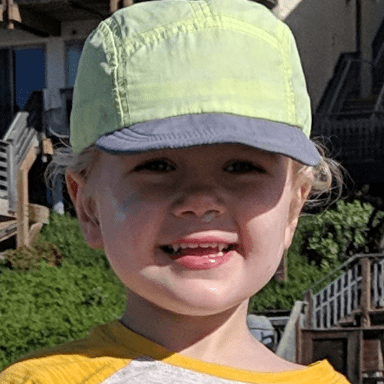}&
  \includegraphics[width=\itemwidth]{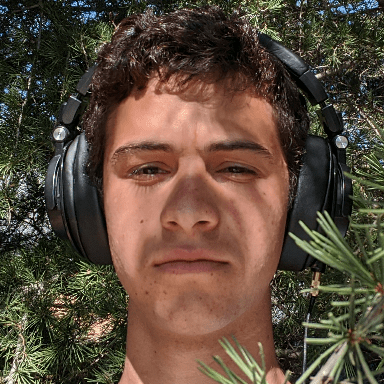}&
  \includegraphics[width=\itemwidth]{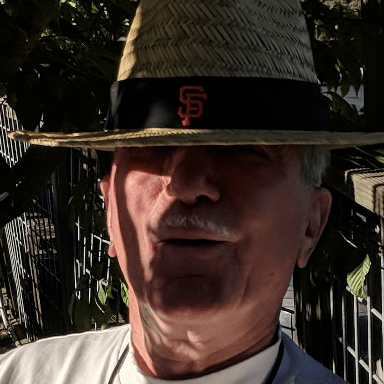}&
  \includegraphics[width=\itemwidth]{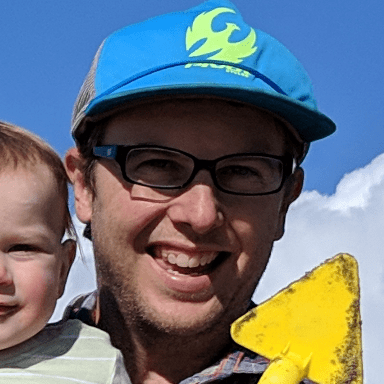}\\
  \includegraphics[width=\itemwidth]{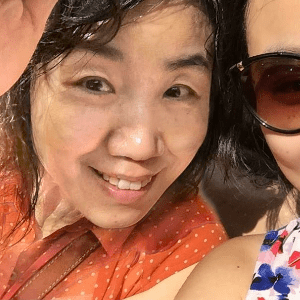}&
  \includegraphics[width=\itemwidth]{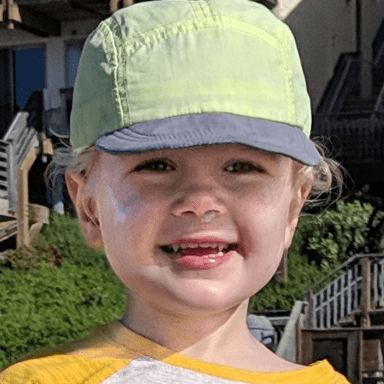}&
  \includegraphics[width=\itemwidth]{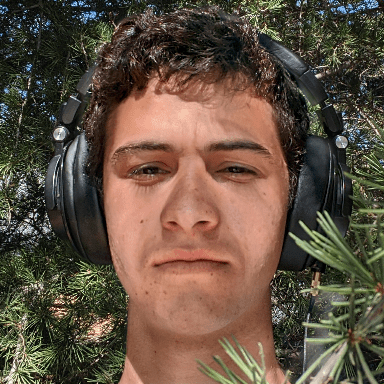}&
  \includegraphics[width=\itemwidth]{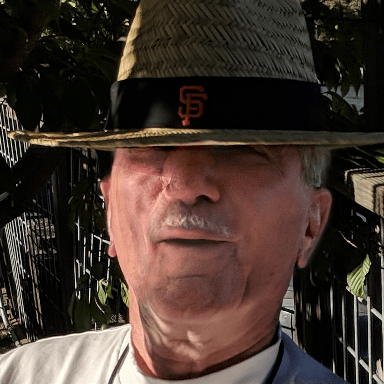}&
  \includegraphics[width=\itemwidth]{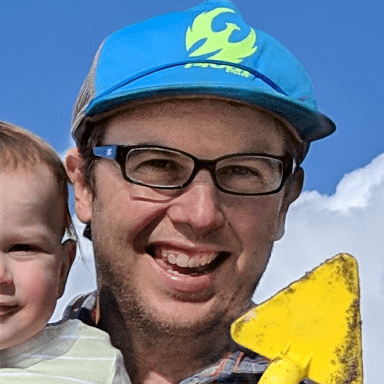}\\
\end{tabular}
\vspace{-0.cm}
\captionof{figure}{Foreign shadow removal results of our model on our \texttt{wild} test dataset. Input images that contain unwanted foreign shadows (top) are processed by our foreign shadow removal model (bottom).
Though real-world foreign shadows exhibit significant variety in terms of shape, softness, and color, our foreign shadow removal model is able to successfully generalize to these challenging real-world images despite having been trained entirely on our synthetic training data (Section~\ref{sec:foreign-syn}).}
\label{fig:exp-foreign-result}
\end{figure*}

We qualitatively evaluate both our foreign shadow removal model and our facial shadow softening model using an additional dataset:
\paragraph{wild}
We collect $100$ ``in the wild'' portrait images of varied human subjects that contain a mix of different foreign and facial shadows. Images are taken from the Helen dataset~\cite{le2012interactive}, the HDRnet dataset~\cite{gharbi2017deep}, and our own captures. These images are processed by our foreign shadow removal model, our facial shadow softening model, or both, to generate enhanced outputs that give a sense of the qualitative properties of both components of our model. See Figures~\ref{fig:teaser}, \ref{fig:exp-foreign-result}, \ref{fig:exp-facial-result}, and the supplement for results.

\begin{figure*}
\begin{tabular}{@{}c@{\,\,}c@{\,\,}c@{\,\,}c@{\,\,}c@{\,\,}c@{}}
  \includegraphics[trim=50 20 0 30,clip,width=1.14in]{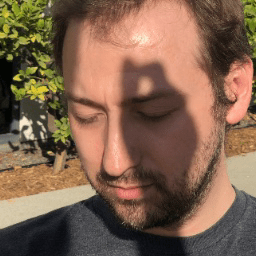}&
  \includegraphics[trim=50 20 0 30,clip,width=1.14in]{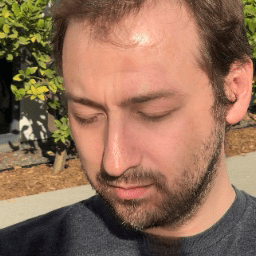}&
  \includegraphics[trim=50 20 0 30,clip,width=1.14in]{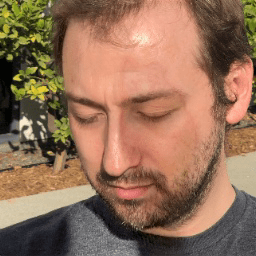}&
  \includegraphics[trim=50 20 0 30,clip,width=1.14in]{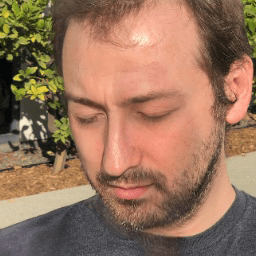}&
  \includegraphics[trim=50 20 0 30,clip,width=1.14in]{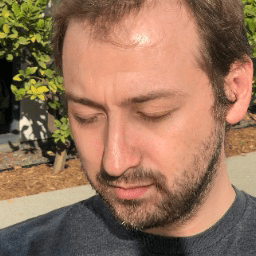}&
  \includegraphics[trim=50 20 0 30,clip,width=1.14in]{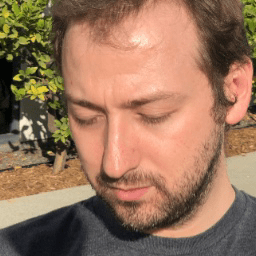}\\
  \includegraphics[trim=50 20 0 30,clip,width=1.14in]{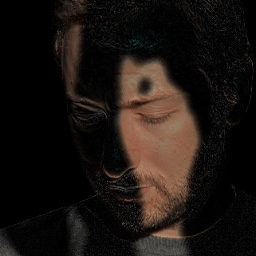}&
  \includegraphics[trim=50 20 0 30,clip,width=1.14in]{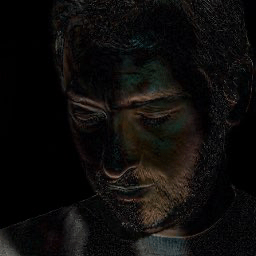}&
  \includegraphics[trim=50 20 0 30,clip,width=1.14in]{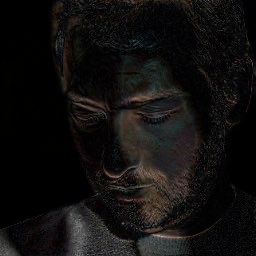}&
  \includegraphics[trim=50 20 0 30,clip,width=1.14in]{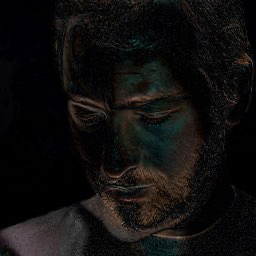}&
  \includegraphics[trim=50 20 0 30,clip,width=1.14in]{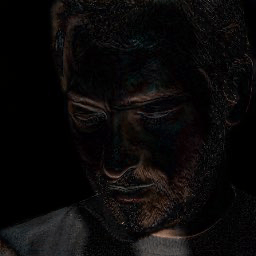}&
  \includegraphics[trim=50 20 0 30,clip,width=1.14in]{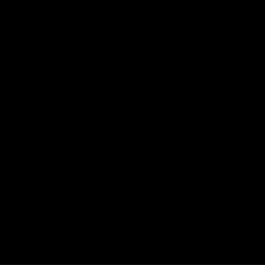}
  \\
  \footnotesize (a) Input Image &
  \footnotesize (b) Our Model, \nosv &
  \footnotesize (c) Our Model, \noss &
  \footnotesize (d) Our Model, \nocolor &
  \footnotesize (e) Our Model &
  \footnotesize (f) Ground Truth \\
\end{tabular}\vspace{-0.2cm}
\captionof{figure}{
A visualization of an ablation study of our foreign shadow removal algorithm
as different aspects of our foreign shadow synthesis model (Section~\ref{sec:foreign-syn}) are removed. 
The ``\nosv'', ``\noss'', and ``\nocolor'' ablations show our model trained on synthesized data \emph{without} modeling spatial variation, approximate subsurface scattering, or color perturbation, respectively.
The top row shows the images generated by each model, and the bottom row shows the difference between each output and the ground truth image (f). Our complete model (e) clearly outperforms the others. Notice the red-colored residual along the shadow edge in the model trained without approximating subsurface scattering (c), and the color inconsistency in the removed region in the model trained without color perturbation (d). A quantitative evaluation on the entire set \texttt{foreign-real} is shown in Table~\ref{tbl:foreign-ablation}.}
\label{fig:exp-ablation}
\end{figure*}

\subsection{Ablation Study of Foreign Shadow Synthesis}
\label{sec:exp-ablation}
Our foreign shadow synthesis technique (Section~\ref{sec:foreign-syn}) simulates the complicated effect of foreign shadows on the appearance of human subjects.
We evaluate this technique by removing each of the three components and measuring model performance.
Our three ablations are: 1) ``\nosv'': synthesis without spatially varying blur or the intensity variation of the shadow, 2) ``\noss'': synthesis where the approximated subsurface scattering of skin has been removed, and 3) ``\nocolor'': synthesis where the color perturbation to generate the shadow image is not randomly changed.
Quantitative results for this ablation study
on our \texttt{foreign-syn} and \texttt{foreign-real} datasets
can be found in Table~\ref{tbl:foreign-ablation}, and qualitative results for a test image from \texttt{foreign-real} are shown in Figure~\ref{fig:exp-ablation}.
\begin{figure}\centering
	\setlength{\tabcolsep}{0.0cm}
	\renewcommand{\arraystretch}{1.}
	\newcommand{\quantTit}[1]{\multicolumn{3}{c}{\scriptsize #1}}
	\newcommand{\quantSec}[1]{\scriptsize #1}
	\newcommand{\quantInd}[1]{\scriptsize #1}
	\newcommand{\quantVal}[1]{\scalebox{0.83}[1.0]{$ #1 $}}
	\newcommand{\quantBes}[1]{\scalebox{0.83}[1.0]{$\uline{ #1 }$}}
	\footnotesize
	\captionof{table}{A quantitative ablation study of our foreign shadow synthesis model in terms of PSNR, SSIM, and LPIPS. Ablating any component of our synthesis model hurts the performance of the resulting model.}
	\begin{tabularx}{\columnwidth}{
	@{\hspace{0.1cm}} X P{1.12cm}
	@{\hspace{-0.2cm}} P{1.12cm}
	@{\hspace{-0.2cm}} P{1.32cm} P{1.12cm} @{\hspace{-0.2cm}} P{1.12cm}
	@{\hspace{-0.2cm}} P{1.12cm}}
		\toprule
			& \quantTit{Rendered Test Set (\texttt{foreign-syn})} & \quantTit{Real Test Set (\texttt{foreign-real})}
		\\ \cmidrule(l{2pt}r{2pt}){2-4} \cmidrule(l{2pt}r{2pt}){5-7}
			Synthesis Model &
			 \quantSec{PSNR}\quantInd{$\uparrow$} & \quantSec{SSIM}\quantInd{$\uparrow$} & \quantSec{LPIPS} \quantInd{$\downarrow$} & \quantSec{PSNR}\quantInd{$\uparrow$} & \quantSec{SSIM}\quantInd{$\uparrow$} & \quantSec{LPIPS}\quantInd{$\downarrow$}
		\\ \midrule
Ours, ``\nocolor'' & \quantVal{26.248} & \quantVal{0.818} & \quantVal{0.079} & \quantVal{21.387} & \quantVal{0.766} & \quantVal{0.085}
\\
Ours, ``\nosv'' & \quantVal{27.546} & \quantVal{0.830} & \quantVal{0.058} & \quantVal{22.095} & \quantBes{0.782} & \quantVal{0.081}
\\
Ours, ``\noss'' & \quantVal{26.996} & \quantVal{0.809} & \quantVal{0.074} & \quantVal{21.663} & \quantVal{0.770} & \quantVal{0.086}
\\
Ours & \quantBes{29.814} & \quantBes{0.926} & \quantBes{0.054} & \quantBes{23.816} & \quantBes{0.782} & \quantBes{0.074}
		\\ \bottomrule
	\end{tabularx}\vspace{0.2cm}
	\vspace{-0.4cm}
	\label{tbl:foreign-ablation}
\end{figure}


\subsection{Foreign Shadow Removal Quality}
\label{sec:exp-foreign-eval}
\begin{figure}\centering
	\setlength{\tabcolsep}{0.0cm}
	\renewcommand{\arraystretch}{1.}
	\newcommand{\quantTit}[1]{\multicolumn{3}{c}{\scriptsize #1}}
	\newcommand{\quantSec}[1]{\scriptsize #1}
	\newcommand{\quantInd}[1]{\scriptsize #1}
	\newcommand{\quantVal}[1]{\scalebox{0.83}[1.0]{$ #1 $}}
	\newcommand{\quantBes}[1]{\scalebox{0.83}[1.0]{$\uline{ #1 }$}}
	\footnotesize
	\captionof{table}{ A quantitative evaluation of our foreign shadow removal model. We compare against baseline methods of \citet{guo2012paired}, \citet{hu2019direction} (SRD) and \citet{cun2019ghostfree} on both synthetic and real test sets. The input image itself is also included as point of reference. In terms of both image-quality (PSNR) and perceptual-quality (SSIM and LPIPS), our model produces better performance on all three metrics with a large margin. Visual comparisons can be seen in Figure~\ref{fig:exp-foreign-eval}.}
	\begin{tabularx}{\columnwidth}{
	@{\hspace{0.1cm}} X P{1.12cm}
	@{\hspace{-0.2cm}} P{1.12cm}
	@{\hspace{-0.2cm}} P{1.32cm} P{1.12cm}
	@{\hspace{-0.2cm}} P{1.12cm}
	@{\hspace{-0.2cm}} P{1.12cm}}
		\toprule
			& \quantTit{Rendered Test Set (\texttt{foreign-syn})} & \quantTit{Real Test Set (\texttt{foreign-real})}
		\\ \cmidrule(l{2pt}r{2pt}){2-4} \cmidrule(l{2pt}r{2pt}){5-7}
			Removal Model &
			 \quantSec{PSNR}\quantInd{$\uparrow$} & \quantSec{SSIM}\quantInd{$\uparrow$} & \quantSec{LPIPS}\quantInd{$\downarrow$} & \quantSec{PSNR}\quantInd{$\uparrow$} & \quantSec{SSIM}\quantInd{$\uparrow$} & \quantSec{LPIPS}\quantInd{$\downarrow$}
		\\ \midrule
Input Image & \quantVal{20.657} & \quantVal{0.807} & \quantVal{0.206} & \quantVal{19.671} & \quantVal{0.766} & \quantVal{0.115}
\\
\cite{guo2012paired} & \quantVal{19.170} & \quantVal{0.699} & \quantVal{0.359} & \quantVal{15.939} & \quantVal{0.593} & \quantVal{0.269}
\\
\cite{hu2019direction} & \quantVal{20.895} & \quantVal{0.742} & \quantVal{0.238} & \quantVal{18.956} & \quantVal{0.699} & \quantVal{0.148}
\\
\cite{cun2019ghostfree} & \quantVal{22.405} & \quantVal{0.845} & \quantVal{0.173} & \quantVal{19.386} & \quantVal{0.722} & \quantVal{0.133}
\\
Ours & \quantBes{29.814} & \quantBes{0.926} & \quantBes{0.054} & \quantBes{23.816} & \quantBes{0.782} & \quantBes{0.074}
		\\ \bottomrule
	\end{tabularx}\vspace{0.2cm}
	\vspace{-0.4cm}
	\label{tbl:foreign-quant}
\end{figure}
\begin{figure}\centering
	\setlength{\tabcolsep}{0.0cm}
	\renewcommand{\arraystretch}{1.}
	\newcommand{\quantTit}[1]{\multicolumn{3}{c}{\scriptsize #1}}
	\newcommand{\quantSec}[1]{\scriptsize #1}
	\newcommand{\quantInd}[1]{\scriptsize #1}
	\newcommand{\quantVal}[1]{\scalebox{0.83}[1.0]{$ #1 $}}
	\newcommand{\quantBes}[1]{\scalebox{0.83}[1.0]{$\uline{ #1 }$}}
	\footnotesize
	\captionof{table}{A comparison of our facial shadow reduction model against the PR-net of \citet{Sun2019} and an ablation of our model with symmetry. in terms of PSNR, SSIM, and LPIPS on the ``\texttt{facial-syn}'' test dataset. We see that PR-net performs poorly on images that contain harsh facial shadows, and removing the concatenated ``mirrored'' input during training (\ie setting $I_\mathrm{in} = I$) lowers accuracy by all three metrics.}
	\begin{tabularx}{0.8\columnwidth}{
	@{\hspace{0.1cm}} X P{1.42cm}
	@{\hspace{-0.2cm}} P{1.42cm}
	@{\hspace{-0.2cm}} P{1.42cm}}
		\toprule
			Shadow Reduction Model &
			 \quantSec{PSNR}\quantInd{$\uparrow$} & \quantSec{SSIM}\quantInd{$\uparrow$} & \quantSec{LPIPS}\quantInd{$\downarrow$}\\
		\midrule
		    PR-net~\cite{Sun2019} & \quantVal{21.639} & \quantVal{0.709} & \quantVal{0.152}\\
            Ours w/o Symmetry & \quantVal{24.232} & \quantVal{0.826} & \quantVal{0.065}\\
            Ours & \quantBes{26.740} & \quantBes{0.914} & \quantBes{0.054}\\
        \bottomrule
	\end{tabularx}\vspace{0.2cm}
	\vspace{-0.4cm}
	\label{tbl:facial-quant}
\end{figure}
Because no prior work appears to address the task of foreign shadow removal for human faces, we compare our model against general-purpose shadow removal methods: the state-of-the-art learning-based method of \citet{cun2019ghostfree}\footnote{https://github.com/vinthony/ghost-free-shadow-removal} that uses a generative model to synthesize and then remove shadows, a customized network with attention mechanism designed by \citet{hu2019direction}\footnote{https://github.com/xw-hu/DSC} for shadow detection and removal, and the non-learning-based method of \citet{guo2012paired} that relies on image segmentation and graph cuts. The original implementation from~\cite{guo2012paired} is not available publicly, so we use a reimplementation\footnote{https://github.com/kittenish/Image-Shadow-Detection-and-Removal} that is able to reproduce the results of the original paper. We use the default parameters settings for this code, as we find that tuning its parameters did not improve performance for our task. \citet{hu2019direction} provide two models trained on two general-purpose shadow removal benchmark datasets (SRD and ISTD), we use the SRD model as it performs better than the ISTD model on our evaluation dataset.

\begin{figure*}
\setlength{\tabcolsep}{0.1cm}
\setlength{\itemwidth}{2.8cm}
\begin{tabular}{cccccc}
  \includegraphics[width=\itemwidth]{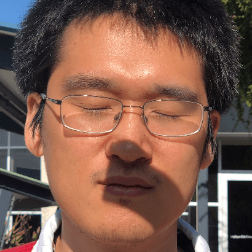}&
  \includegraphics[width=\itemwidth]{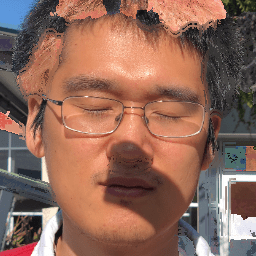}&
  \includegraphics[width=\itemwidth]{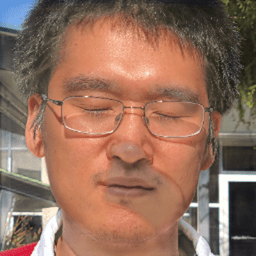}&
  \includegraphics[width=\itemwidth]{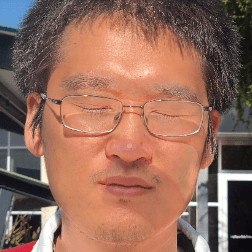}&
  \includegraphics[width=\itemwidth]{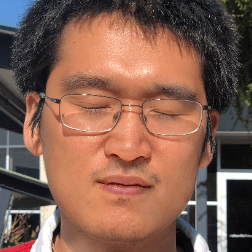}&
  \includegraphics[width=\itemwidth]{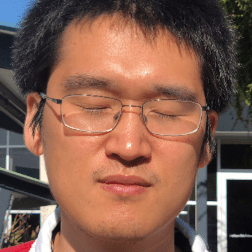} \\
  \includegraphics[width=\itemwidth]{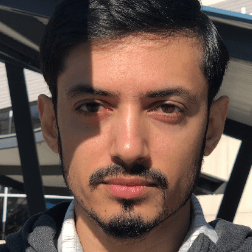}&
  \includegraphics[width=\itemwidth]{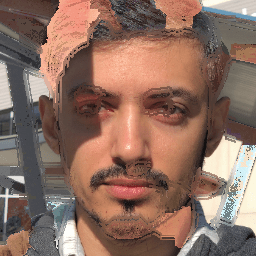}&
  \includegraphics[width=\itemwidth]{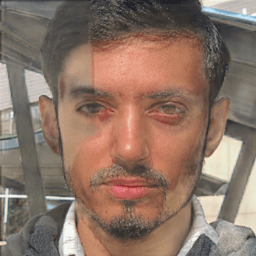}&
  \includegraphics[width=\itemwidth]{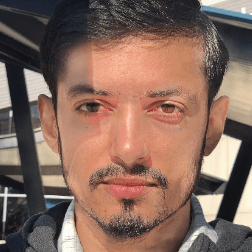}&
  \includegraphics[width=\itemwidth]{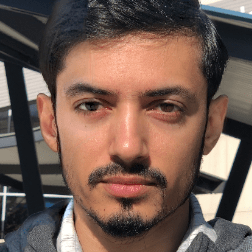}&
  \includegraphics[width=\itemwidth]{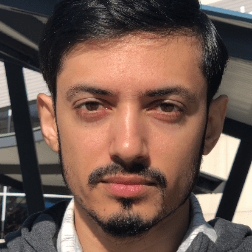} \\
  \includegraphics[width=\itemwidth]{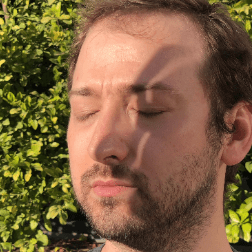}&
  \includegraphics[width=\itemwidth]{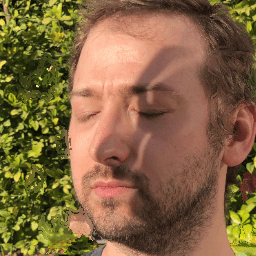}&
  \includegraphics[width=\itemwidth]{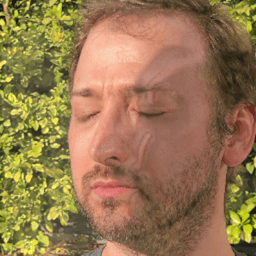}&
  \includegraphics[width=\itemwidth]{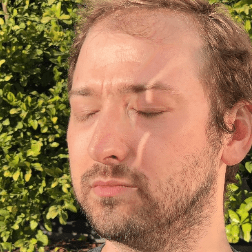}&
  \includegraphics[width=\itemwidth]{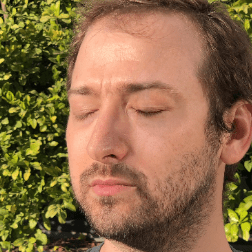}&
  \includegraphics[width=\itemwidth]{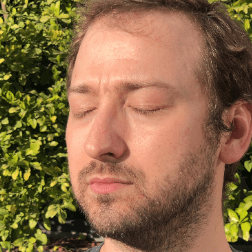} \\
    \small (a) Input Image &
    \small (b) \citet{guo2012paired} &
    \small (c) \citet{hu2019direction} &
    \small (d) \citet{cun2019ghostfree} &
    \small (e) Our Model &
    \small (f) Ground Truth \\
\end{tabular}
\vspace{-0.cm}
\captionof{figure}{Foreign shadow removal results on images from our \texttt{foreign-real} test dataset. The method of ~\citet{guo2012paired} often incorrectly identifies dark image regions as shadows and removes them, while also failing to identify real shadows (b).
The deep learning approaches of \citet{cun2019ghostfree} and \citet{hu2019direction} (c, d) do a better job of correctly identifying shadow regions but often fail to remove shadows completely, and also change the overall brightness and tone of the image in a way that does not preserve the authenticity of the input image.
Our method is able to entirely remove foreign shadows while still preserving the overall appearance of the subject (e), thereby producing output images that more closely resemble the ground truth (f).
}
\label{fig:exp-foreign-eval}
\end{figure*}

We evaluate these baseline methods on our \texttt{foreign-syn} and \texttt{foreign-real} datasets, as these both contain ground truth shadow-free images. We compute PSNR, SSIM~\cite{Wang_TIP_2004} and a learned perceptual metric LPIPS~\cite{Rizhang_CVPR_2018} between the ground truth and the output. Results are shown in Table~\ref{tbl:foreign-quant} and Figure~\ref{fig:exp-foreign-eval}. Our model outperforms these baselines by a large margin. Please see video at 03:03 for more results.

\begin{figure*}
\begin{center}
\includegraphics[width=0.96\linewidth]{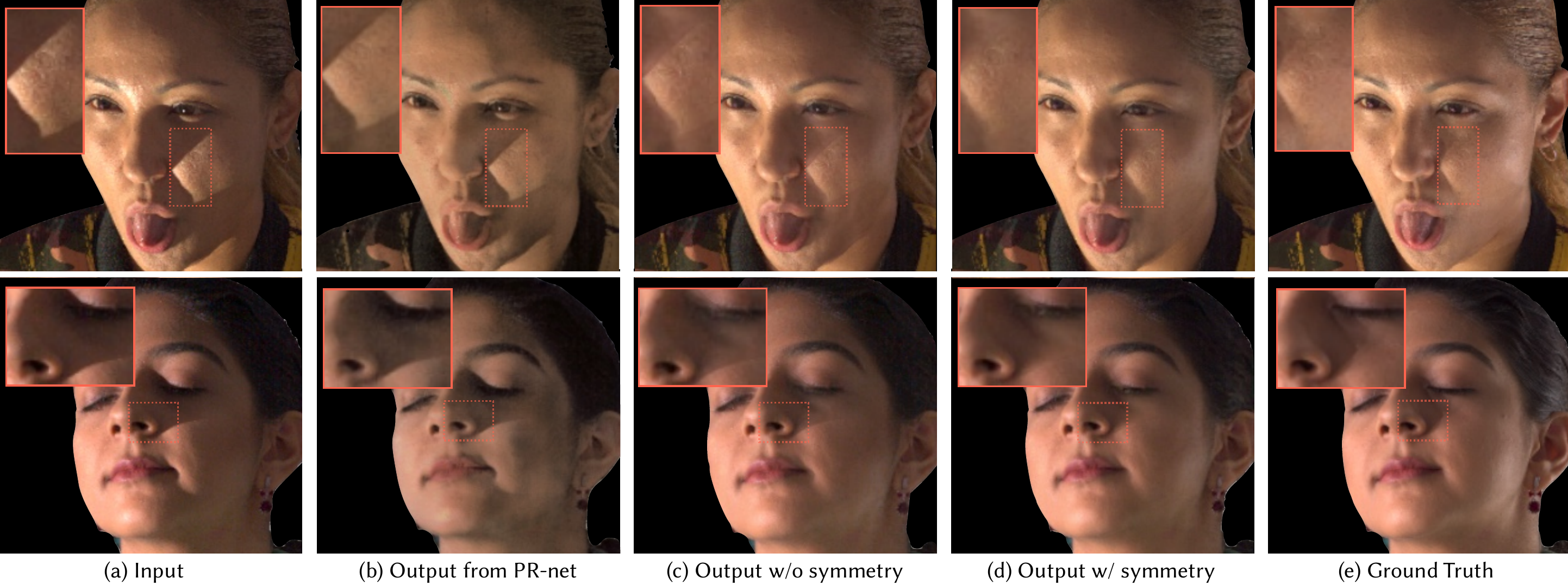}\\
\vspace{-0.3cm}
\captionof{figure}{Facial shadow softening results on \texttt{facial-syn}. We compare against the portrait relighting model (PR-net)~\citet{Sun2019} by applying a blur to its estimated environment light and relighting the input image with that blurred environment map. PR-net is able to successfully soften low frequency shadows but struggles with harsh facial shadows (b).
The ablation of our model without our symmetry component (Section~\ref{sec:facial-sym}) also underperforms on these harsh facial shadows (c).
Our complete model successfully softens these hard shadows (d), as it is able to reason about the bilateral symmetry of the subject and ``borrow'' pixels with similar reflectance and geometry from the opposite side of the face.
}
\label{fig:exp-ablation-sym}
\vspace{-0.1cm}
\end{center}
\end{figure*}

\begin{figure*}
\setlength{\tabcolsep}{0.08cm}
\setlength{\itemwidth}{3.25cm}
\begin{tabular}{ccccc}
  \includegraphics[width=\itemwidth]{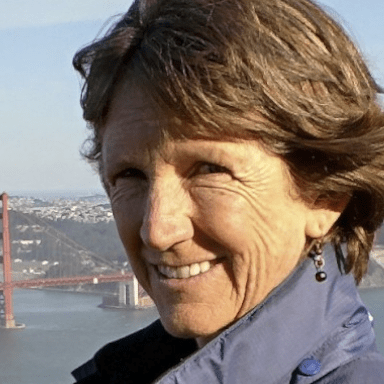}&
  \includegraphics[width=\itemwidth]{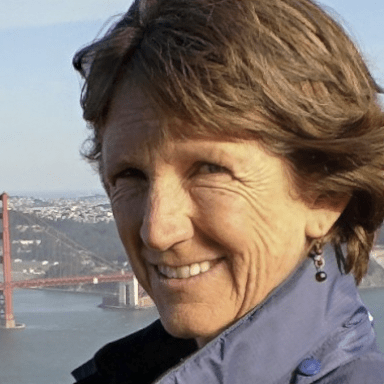}&
  \includegraphics[width=\itemwidth]{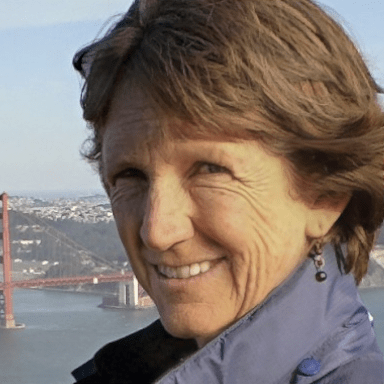}&
  \includegraphics[width=\itemwidth]{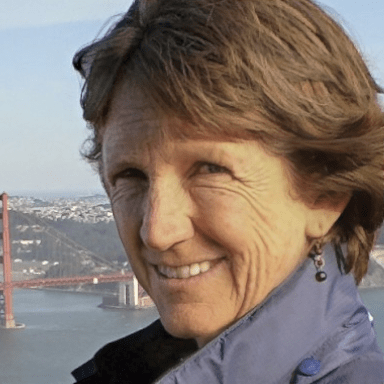}&
  \includegraphics[width=\itemwidth]{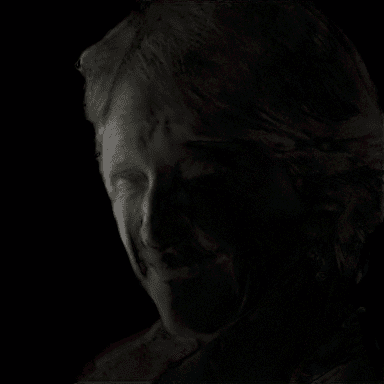} \\
  \includegraphics[width=\itemwidth]{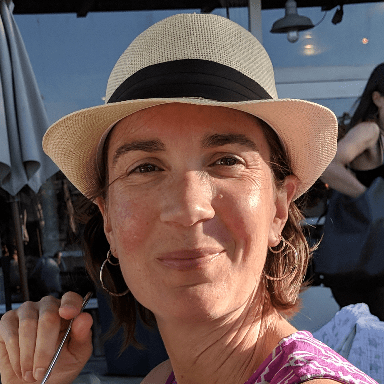}&
  \includegraphics[width=\itemwidth]{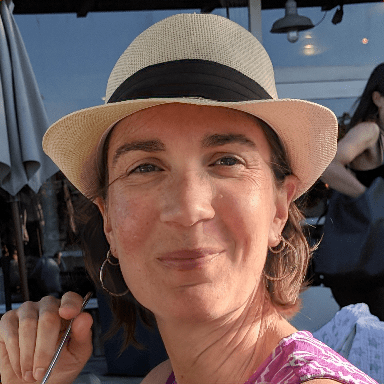}&
  \includegraphics[width=\itemwidth]{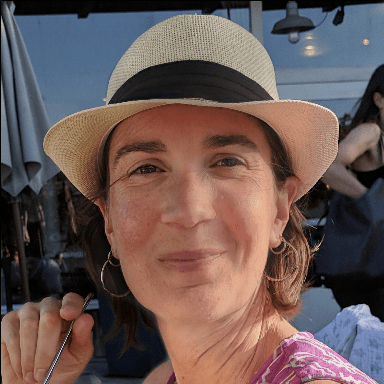}&
  \includegraphics[width=\itemwidth]{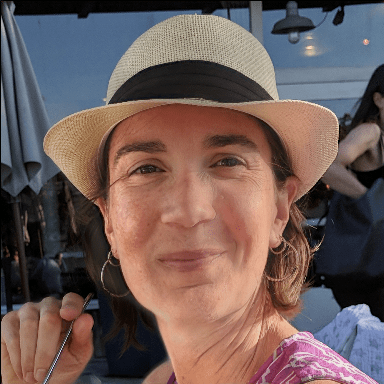}&
  \includegraphics[width=\itemwidth]{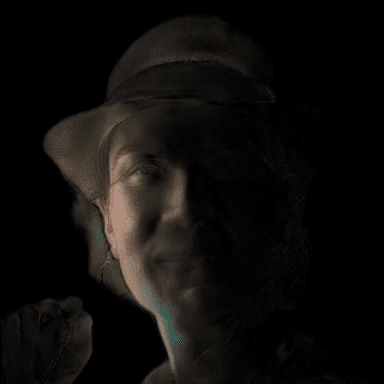} \\
  \includegraphics[width=\itemwidth]{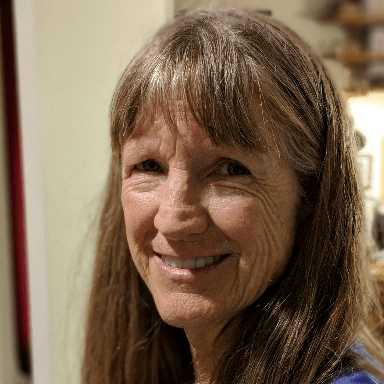}&
  \includegraphics[width=\itemwidth]{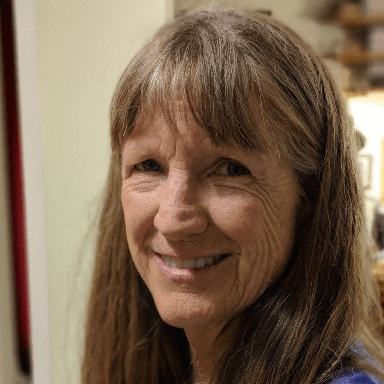}&
  \includegraphics[width=\itemwidth]{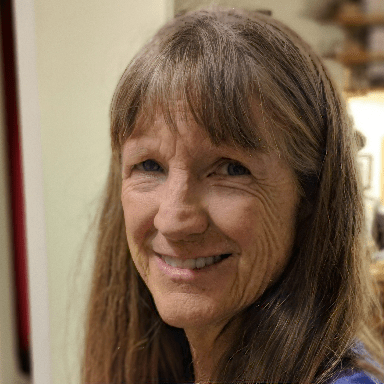}&
  \includegraphics[width=\itemwidth]{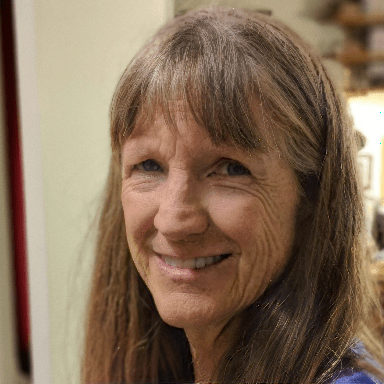}&
  \includegraphics[width=\itemwidth]{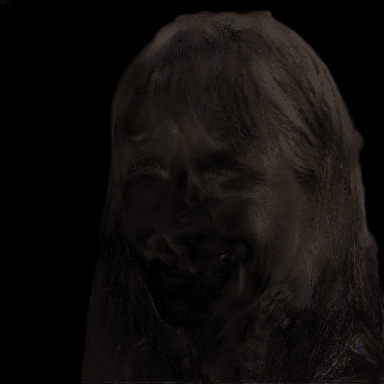} \\
    \small (a) Input &
    \small (b) Output ($\lightcount = 10$) &
    \small (c) Output ($\lightcount = 40$) &
    \small (d) Output &
    \small (e) |(c) - (d)| \\
    & & &
    \small ($\lightcount = 40$, $\fillintensity=\fillintensity^{\operatorname{max}}$) &
\end{tabular}
\vspace{-0.25cm}
\captionof{figure}{Facial shadow softening results on images from \texttt{wild}. Input images may contain harsh facial shadows, such as around the eyes (row 1) and by the subject's cheek (row 3).
Applying our facial shadow softening model with a variable ``light size'' $\lightcount$ produces images with softer shadows (b, c). The specular reflection also gets suppressed, which is a desired photographic practice as specular highlights are often distracting and obscuring the surface of the subject.
Additionally, the lighting ratio component of our model
reduces the contrast induced by facial shadows (d) by adding a synthetic fill light with intensity $\fillintensity$, set here to
the maximum value used in training
(Section ~\ref{sec:facial-syn}), in the direction opposite to the detected key light, as visualized in (e).
}
\label{fig:exp-facial-result}
\end{figure*}

\subsection{Facial Shadow Softening Quality}
\label{sec:exp-facial-eval}
Transforming harsh facial shadows to soft in image space is roughly equivalent to relighting a face with a blurred version of the dominant light source in the original lighting environment.
We compare our facial softening model against the portrait relighting method from~\citet{Sun2019}, by applying a Gaussian blur to the estimated environment map from the model and then pass to the decoder for relighting. The amount of blur to apply, however, cannot map exactly to our light size parameter. We experiment with a few blur kernel values and choose the one that produces the minimum mean pixel error with the ground truth. We do this for each image, and show qualitative comparisons in Figure~\ref{fig:exp-facial-result}. In Table~\ref{tbl:facial-quant}, we compare our model against the \citet{Sun2019} baseline and against an ablation of our model without symmetry, and demonstrate an improvement with respect to both. For all comparisons, we use \texttt{facial-syn}, which has ground truth soft facial shadows. Please see video at 03:35 for more results.

\subsection{Preprocessing for Portrait Relighting}
\begin{figure}
\setlength{\tabcolsep}{0.1cm}
\setlength{\itemwidth}{3.6cm}
\begin{tabular}{cc}
  \includegraphics[width=\itemwidth]{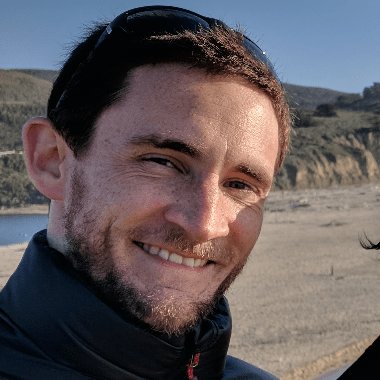}&
  \includegraphics[width=\itemwidth]{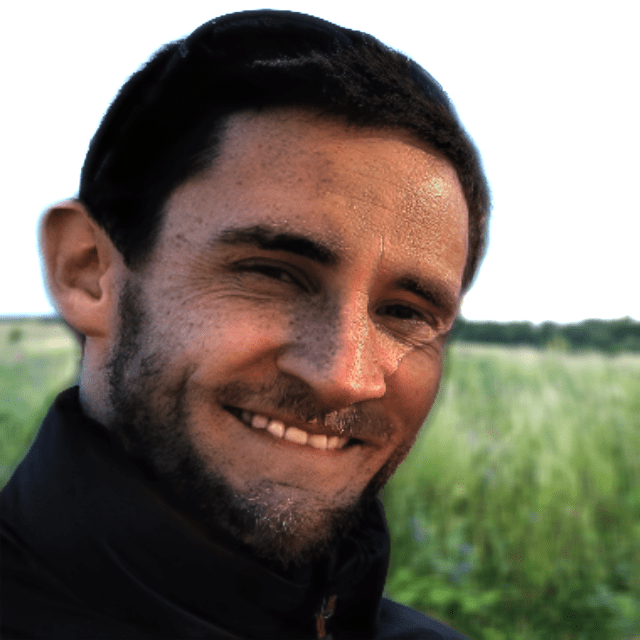} \\
  \footnotesize (a) Input &
  \footnotesize (b) \citet{Sun2019} applied to (a) \\
  \includegraphics[width=\itemwidth]{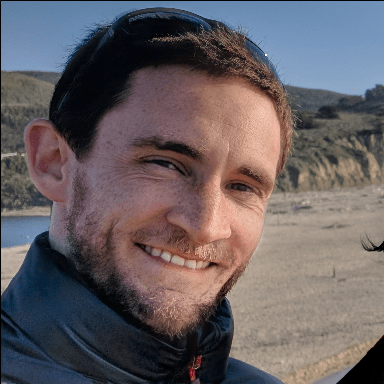} &
  \includegraphics[width=\itemwidth]{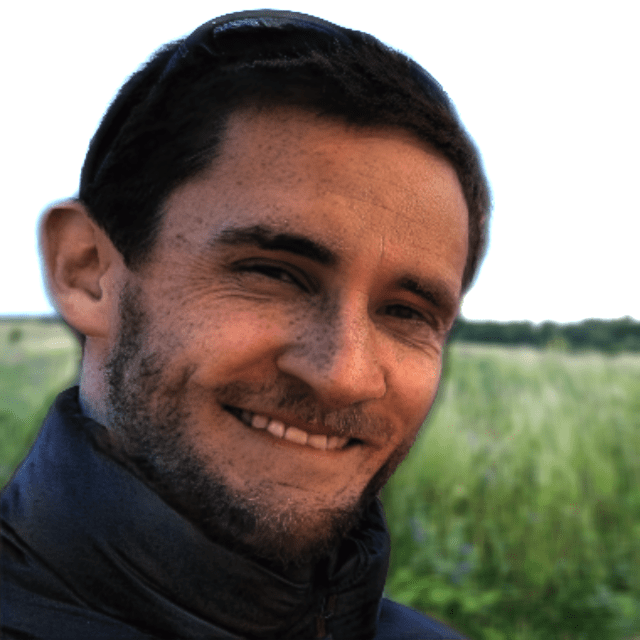}\\
  \footnotesize (c) Our output &
  \footnotesize (d) \citet{Sun2019} applied to (c) \\
\end{tabular}
\vspace{-0.2cm}
\captionof{figure}{The portrait relighting technique of \citet{Sun2019}
provides an alternative approach for shadow manipulation.
However, applying this technique to input images that contain foreign shadows and harsh facial shadows (a) often results in relit images in which these foreign and facial shadows persist as artifacts (b).
If this same portrait relighting technique is instead applied to the output images of our model (c), it produces a less jarring (though still somewhat suboptimal) rendering of the subject (d).
}
\label{fig:exp-relight}
\end{figure}

Our method can also be used as a ``preprocessing'' step for image modification algorithms such as portrait relighting~\cite{Sun2019, zhou2019deep}, which modify or replace the illumination of the input image. 
Though often effective, these portrait relighting techniques sometimes produce suboptimal renderings when presented with input images that contain foreign shadows or harsh facial shadows.
Our technique can improve a portrait relighting solution: our model can be used to remove these unwanted shadowing effects, producing a rendering that can then be used as input to a portrait relighting solution, resulting in an improved final rendering.
See Figure~\ref{fig:exp-relight} for an example.

\begin{figure}
\setlength{\itemwidth}{0.44\linewidth}
\begin{center}
\begin{tabular}{cc}
  \includegraphics[width=\itemwidth]{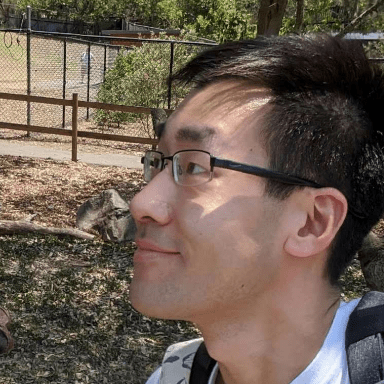}&
  \includegraphics[width=\itemwidth]{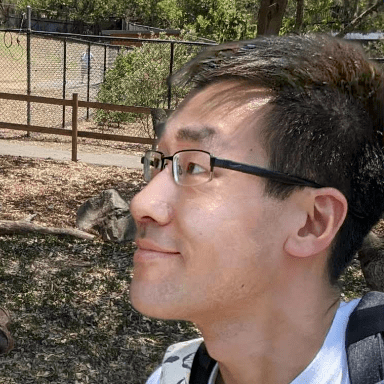}\\
  \multicolumn{2}{c}{\footnotesize (a) Some fine-detailed shadows (cast by hair) is visible in the output.}\\
  \includegraphics[width=\itemwidth]{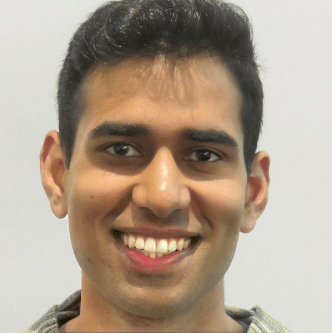}&
  \includegraphics[width=\itemwidth]{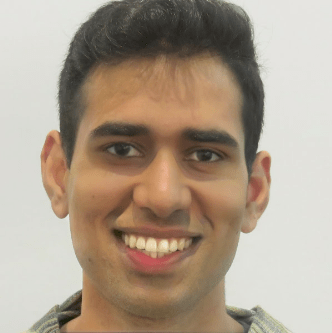}\\
  \multicolumn{2}{c}{\footnotesize (b) Highly symmetric facial shadows are less likely to be softened.}\\
  \includegraphics[width=\itemwidth]{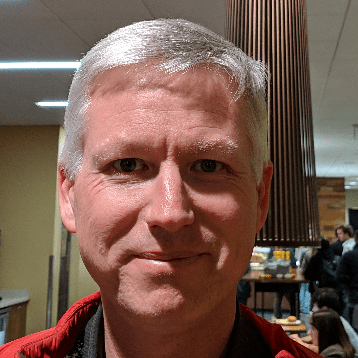}&
  \includegraphics[width=\itemwidth]{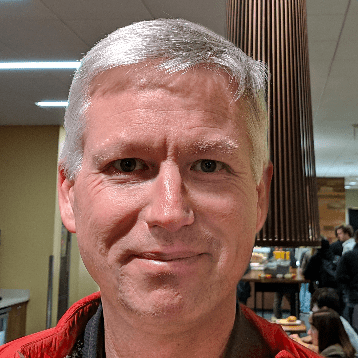}\\
  \multicolumn{2}{c}{\footnotesize (c) Facial shadows may resemble foreign shadows and get removed.}\\
\end{tabular}
\end{center}
\vspace{-0.2cm}
\captionof{figure}{Example failure cases from our \texttt{wild} dataset. We notice limitations of our foreign shadow removal model in handling fine-detailed structures (a), of our facial shadow softening model reducing highly facial shadows (b), and of the models not correctly separating the two types of shadows (c).}
\label{fig:exp-failure}
\end{figure}

\section{Limitations}
\label{sec:limitations}

Our proposed model is not without its limitations, some of which we can identify in our \texttt{wild} dataset.
When foreign shadows contain many finely-detailed structures (which are underrepresented in training), our output may retain visible residuals of those (Figure~\ref{fig:exp-failure}(a)). While exploiting the bilateral symmetry of the subject significantly improves our facial softening model's performance, this causes our model to sometimes fail to remove shadows that also happen to be bilaterally symmetric (Figure~\ref{fig:exp-failure}(b)). Because the training data of our shadow softening model is rendered by increasing the light size---a simple lighting setup that introduces bias towards generating diffused-looking images. For example, when the ``light size'' is set high in Figure~\ref{fig:exp-facial-result} (c), our shadow softening model generates images with a ``flat'' appearance and smooths out high frequency details in the hair regions that could have been preserved if different lighting setups are used for face and hair during training data generation.

Our model assumes that shadows belong to one of two categories (``foreign'' and ``facial'') but these two categories are not always entirely distinct and easily-separated.
Because of this, sufficiently harsh facial shadows may be erroneously detected and removed by our foreign shadow removal model (Figure~\ref{fig:exp-failure}(c)).
This suggests that our model may benefit from a unified approach for both kinds of shadows, though this approach is somewhat at odds with the constraints provided by image formation and our datasets: a unified learning approach would require a unified source of training data, and it is not clear how existing light stage scans or in-the-wild photographs could be used to construct a large, diverse, and photorealistic dataset in which both foreign and facial shadows are present and available as ground-truth.

Constructing a real-world dataset for our task that contains ground-truth is challenging. Though the \texttt{foreign-real} dataset used for qualitatively evaluating our foreign shadow removal algorithm is sufficiently diverse and accurate to evaluate different algorithms, it has some shortcomings. This dataset is not large enough to be used for training, and does not provide a means for evaluating facial shadow softening.
This dataset also assumes that all foreign shadows are cast by a single occluder blocking the light of a single dominant illuminant, while real-world instances of foreign shadows often involve multiple illuminants and occluders.
Additionally, to satisfy our single-illuminant assumption, this dataset had to be captured in real-world environments that have one dominant light source (\eg, outdoors in the midday sun).
This gave us little control over the lighting environment,
and resulted in images with high dynamic ranges and therefore ``deep'' dark shadows, which may degrade (via noise and quantization) image content in shadowed regions. 
A real-world dataset that addresses these issues be invaluable for evaluating and improving portrait shadow manipulation algorithms.

\section{Conclusion}
We have presented a new approach for enhancing casual portrait photographs with respect to lighting and shadows, particularly in terms of foreign shadow removal, facial shadow softening, and lighting ratio balancing.
When synthesizing training data the foreign shadow removal task, we observe the value of in-the-wild images with a shadow synthesis model that accounts for the irregularity of foreign shadows in the real world.
Motivated by the physical tools used by photographers in studio environments, we have demonstrated how Light Stage scans can be used to produce training data for facial shadow softening.
We have presented a mechanism for allowing convolutional neural networks to exploit the inherent bilateral symmetry of human subjects, and we have demonstrated that this improves the performance of facial shadow softening.
Given just a single image of a human subject taken in an unknown and unconstrained environment, our complete system is able to remove unwanted foreign shadows, soften harsh facial shadows, and balance the image's lighting ratio to produce a flattering and realistic portrait image.

\begin{acks}
We thank Augmented Perception from Google for collecting the light stage data. Ren Ng is supported by a Google Research Grant. This work is also supported by a number of individuals from various aspects. We thank Marc Levoy, Kevin (Jiawen) Chen and anonymous reviewers for helpful discussions on the paper. We thank Michael Milne and Andrew Radin for insightful discussions on professional lighting principles and practice. We thank Timothy Brooks for demonstrating Photoshop portrait shadow editing, and to Xiaowei Hu for running the baseline algorithm. We are also grateful to people who kindly consent to be in the video and result images.
\end{acks}

\bibliography{ref} 
\bibliographystyle{ACM-Reference-Format}

\end{document}